\pdfoutput=1
\documentclass[11pt]{article}
\usepackage[letterpaper,margin=0.9in]{geometry}
\usepackage[T1]{fontenc}
\usepackage[utf8]{inputenc}
\usepackage{lmodern}
\usepackage{textcomp}
\usepackage{amsmath,amssymb}
\usepackage{graphicx}
\usepackage{booktabs}
\usepackage{array}
\usepackage{tabularx}
\usepackage{longtable}
\usepackage{xltabular}
\usepackage{xcolor}
\usepackage{tikz}
\usetikzlibrary{positioning,arrows.meta,calc}
\usepackage[font=small,labelfont=bf,labelsep=period]{caption}
\usepackage[section]{placeins}
\usepackage{microtype}
\usepackage{cite}
\usepackage[hidelinks]{hyperref}
\hypersetup{pdfauthor={Joyjeet Singh},pdftitle={Auditing a Learned Simulator for Legged Robots: The Robotic World Model on ANYmal D}}
\providecommand{\tightlist}{\setlength{\itemsep}{0pt}\setlength{\parskip}{0pt}}

\DeclareUnicodeCharacter{2212}{\textminus}
\definecolor{accent}{HTML}{2A78D6}
\definecolor{edge}{HTML}{898781}
\definecolor{quiet}{HTML}{52514E}

\title{Auditing a Learned Simulator for Legged Robots: The Robotic World Model on ANYmal D}
\author{Joyjeet Singh\\ \small Independent Researcher\\ \small Chandigarh, India\\ \small \texttt{joyjeetsingh1@gmail.com}}
\date{}

\begin{document}
\maketitle

\begin{abstract}
A legged robot can learn to walk inside a \emph{learned simulator}: a neural network that predicts the robot's next state and is then run forward on its own predictions. Such simulators are usually trained and checked one step ahead, although a policy uses them for a hundred steps or more. We audit one published system from end to end, without a GPU and mostly on a 2017 laptop: the dynamics model of the Robotic World Model (RWM) for the ANYmal D quadruped, and the uncertainty estimate of its follow-up, RWM-U.

We rebuilt the model and matched the released code's outputs, losses and gradients exactly before training anything. We then measured 101 trained simulators, 200 compressed copies and 38 simulators of three other architectures. Finally, we trained policies inside simulators and tested them in a MuJoCo model of the robot that we built by hand. Five results follow.

\begin{enumerate}
\def\labelenumi{\arabic{enumi}.}
\tightlist
\item
  \textbf{RWM's central claim holds, and reverses at one step.} With each of ten episodes held out once, training on the model's own rollouts beat one-step training at 368 steps on every episode (registered). One step ahead, the one-step-trained model was better.
\item
  \textbf{One-step checks therefore misjudge design choices.} Teacher forcing and one-step training looked better one step ahead, but were 2.0 and 2.6 times as wrong at 2 s. Pruning 30\% of the weights kept 90\% of the model's skill one step ahead and 39\% at 2 s; a short fine-tuning restored it (registered).
\item
  \textbf{The uncertainty ranks errors but mis-sizes them.} Ensemble disagreement ranked rollout error well (Spearman +0.90 at 100 steps), yet covered 4.6\% of outcomes at one sigma, against 68\% for a calibrated estimate. Each network's own variance collapses, because the implemented loss is smallest at zero variance.
\item
  \textbf{Checks that need no recorded future help, within limits.} An error growth rate and a kinematic residual ranked some pools of simulators well. Neither was shown better than error measured eight steps ahead in a primary test, and neither worked as a run-time alarm.
\item
  \textbf{At the robot level, accuracy has not predicted walking.} Walking policies came only from the released model, its two lightest compressed copies, and copies of it fine-tuned very briefly on data from our physics model. Twenty fine-tuning steps at the original learning rate, or 160 at a tenth of it, were enough to stop the released model giving walking policies. Every simulator we trained from scratch gave policies that fell or stood still, including one that predicts the test physics more accurately one step ahead. All 18 fine-tuned copies we tested predicted better than the released model one and eight steps ahead, yet 13 of them gave failing policies. Registered tests of whether any check ranks simulators by their policies were not settled, on 9 simulators or on 19. In a secondary test, running the authors' policy inside each simulator ranked them well.
\end{enumerate}

We also report nine of our own numbered claims that we withdrew, and what caught them. Scope: one robot, one trotting gait on flat ground, simulated data, a physics model we built ourselves, and no hardware.
\end{abstract}

\section{Introduction}

Walking controllers for legged robots are usually trained in a physics simulator and then moved to the robot. A newer route replaces the physics simulator with a \emph{learned simulator}: a neural network that predicts the robot's next state from its recent states and its \emph{joint commands} (the 12 joint position targets sent at every control step). The policy is then trained on imagined motion~\cite{r4,r5}. The Robotic World Model (RWM) does this for the ANYmal D quadruped~\cite{r1}. Its follow-up, RWM-U, does it from recorded data alone, using an uncertainty estimate to keep the policy away from states the model does not know~\cite{r2}.

A learned simulator is run forward on its own predictions. Such a run is a \emph{rollout}, and its length is its \emph{horizon}. Each prediction becomes the next input, so a small error can grow~\cite{r5,r6,r7,r9}. RWM trains its model on its own 8-step rollouts, and both RWM and RWM-U imagine 100 steps per policy update, which is 2 s at the robot's 50 Hz control rate~\cite{r1,r2}. The released evaluation code runs the model for 368 steps, or 7.4 s.

A roboticist who wants to use such a model has practical questions, and most of them are normally answered one step ahead:

\begin{itemize}
\tightlist
\item
  Does the published training recipe matter, and at which horizon?
\item
  Can the model's uncertainty be read as a distance to the truth, or only as a ranking?
\item
  Which check should be run before a policy is trained inside the model?
\item
  Can the model be made smaller and faster for a computer on the robot?
\item
  Does any of this predict whether the policy walks?
\end{itemize}

We answer them for one published system, as an independent party with small means. No GPU was used anywhere: models and policies were trained on a 2017 dual-core laptop, and some analysis-only studies ran in a two-core cloud workspace. We rebuilt RWM's dynamics model from the papers and the released code, and showed that the rebuild equals the release in outputs, losses and gradients before training anything. We then trained and measured several hundred simulators on the ten recorded episodes that the authors released. Finally, we built a MuJoCo model of the robot by hand and trained policies inside simulators with the authors' own policy-training code.

\begin{figure}[tbp]
\centering
\begin{tikzpicture}[font=\footnotesize,
  stage/.style={draw=edge, rounded corners=3pt, line width=0.6pt, minimum width=3.45cm,
                minimum height=1.55cm, align=center, inner sep=3pt},
  conn/.style={draw=edge, line width=0.6pt},
  arr/.style={conn, -{Stealth[length=1.8mm,width=1.7mm]}},
  node distance=0.5cm]
\node[stage] (data) {{\small\bfseries Recorded data}\\[2pt]10 episodes, 200 s\\ANYmal D, 50 Hz};
\node[stage, right=of data] (sim) {{\small\bfseries Learned simulator}\\[2pt]next-state network,\\run on its own output};
\node[stage, right=of sim] (pol) {{\small\bfseries Policy training}\\[2pt]inside the simulator,\\100 imagined steps};
\node[stage, right=of pol] (phys) {{\small\bfseries Physics test}\\[2pt]MuJoCo ANYmal D:\\does the policy walk?};
\draw[arr] (data) -- (sim);
\draw[arr] (sim) -- (pol);
\draw[arr] (pol) -- (phys);
\node[draw=edge, rounded corners=3pt, line width=0.6pt, anchor=north west, inner sep=5pt]
  (m1) at ([yshift=-0.75cm]data.south west)
  {\parbox[t][2.0cm][t]{6.55cm}{{\small\bfseries Measured on the simulator alone}\\[3pt]
   error at 1, 8, 100 and 368 steps (Section 5)\\
   ensemble disagreement (Section 6)\\
   growth rate, kinematic residual (Section 7)\\
   the same after compression (Section 8)}};
\node[draw=accent, fill=accent!10, rounded corners=3pt, line width=1pt, anchor=north east, inner sep=5pt]
  (m2) at ([yshift=-0.75cm]phys.south east)
  {\parbox[t][2.0cm][t]{4.9cm}{{\small\bfseries Measured on the policy}\\[3pt]
   fall rate in physics (Section 9)\\
   walking score in physics (Section 9)}};
\draw[conn] (sim.south) -- (sim.south |- m1.north);
\draw[conn] (phys.south) -- (phys.south |- m2.north);
\draw[arr, draw=accent, dashed] (m1.east) -- node[above, text=quiet] {predicts?} (m1.east -| m2.west);
\end{tikzpicture}
\caption{Checks are run on the simulator; what matters is measured on the policy. The stages this paper follows. Sections 4 to 8 measure the simulator on recorded data. Section 9 measures policies trained inside it, in a physics model. Whether the first kind of measurement predicts the second is the question of Sections 7 and 9.}\label{fig:1}
\end{figure}
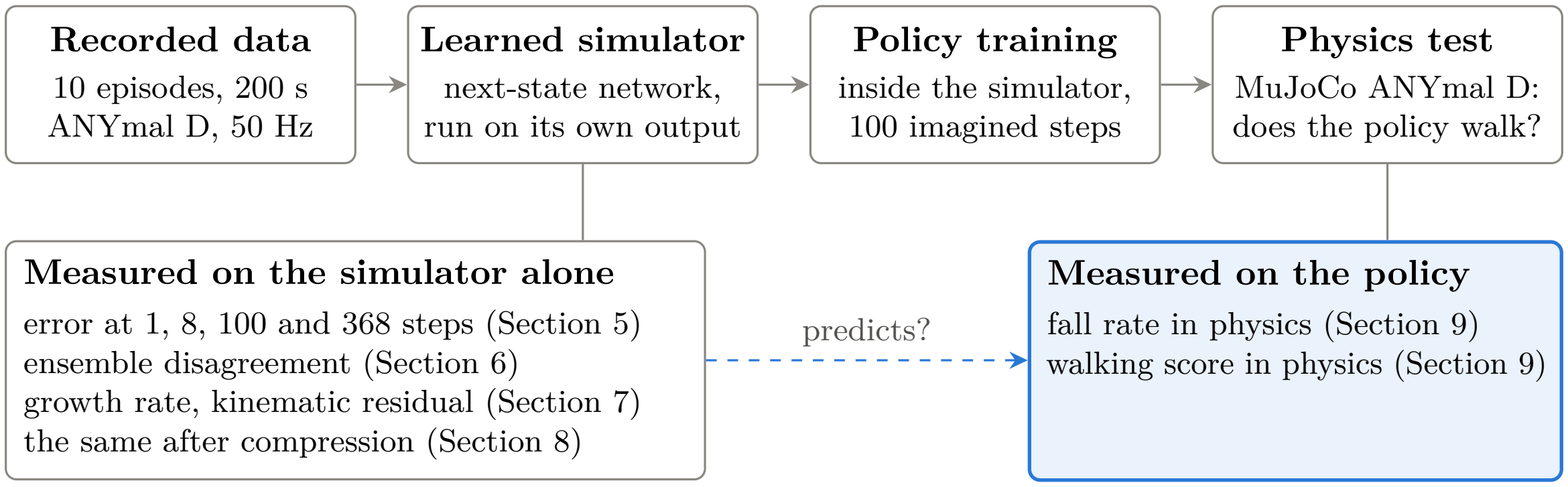

\textbf{Contributions.}

\begin{enumerate}
\def\labelenumi{\arabic{enumi}.}
\tightlist
\item
  \textbf{A verified rebuild and an audit of the release} (Section 4): exact agreement with the released module, two defects in the released pipeline with their measured cost, and the places where the code departs from the papers.
\item
  \textbf{A held-out test of RWM's central training claim} (Section 5): rollout training beats one-step training at long horizons on all ten episodes, the order reverses one step ahead, and the training horizon sets whether a simulator amplifies or damps its own errors.
\item
  \textbf{A measurement of RWM-U's uncertainty} (Section 6): it ranks error well and sizes it wrongly, for three separable reasons; the per-network variance collapses by construction; and a per-horizon rescaling repairs part of it.
\item
  \textbf{A registered comparison of pre-deployment checks} (Section 7): one-step error, 8-step error, an error growth rate and a kinematic residual, across 97 and 255 recurrent simulators, 38 simulators of other architectures and two sources of joint commands, with negative results on run-time alarms.
\item
  \textbf{Compression judged at the horizon of use} (Section 8): quantisation, pruning and fine-tuning scored at every horizon, with measured cost on a laptop CPU.
\item
  \textbf{Robot-level evidence} (Section 9): a hand-built, calibrated MuJoCo model of ANYmal D; policies trained inside 33 simulators; an analysis of why most of them fail; and a registered ladder of 18 fine-tuned copies of the released model. The ladder shows how little fine-tuning on new physics data stops the model giving walking policies, and tests which checks rank such copies by their policies.
\item
  \textbf{A record of what we got wrong} (Section 10): nine withdrawn claims, the measurement mistakes behind them, and the tooling that caught them.
\end{enumerate}

\textbf{What this paper does not show.} Nothing here was run on a robot. The recorded data are simulated, the physics model is ours and not the authors', and everything concerns one robot, one trotting gait and flat ground. We state these limits wherever a result depends on them.

Section 2 places the work. Section 3 describes the robot, the data and how we measure. Sections 4 to 9 give the results. Section 10 lists our own retractions, Section 11 turns the results into working rules, and Sections 12 and 13 give limitations and conclusions.

\section{Background and related work}

\textbf{The system we audit.} RWM learns a recurrent dynamics model of ANYmal D from the robot's own body measurements and uses it as a simulator for policy learning~\cite{r1}. The model reads a history of 32 steps and is trained to predict 8 steps ahead on its own outputs. The paper reports pretraining on 6 million transitions, and policy training with 100 imagined steps per update in 4,096 imagined environments. Its ablation states that the rollout-trained model outperforms a teacher-forced one. RWM-U adds an ensemble of five predictors and subtracts their disagreement from the reward, so that policies can be trained from recorded data without a physics simulator and transferred to hardware~\cite{r2}. The authors released a lightweight code base, one recorded data file and one trained checkpoint. Those three things are all we start from.

\textbf{Learning to walk in simulation.} Legged controllers are commonly trained by reinforcement learning in fast physics simulators and transferred to hardware~\cite{r47,r48,r49}. For ANYmal, part of the gap between simulation and robot is closed by a learned model of the actuators~\cite{r45,r46}. A learned simulator moves the whole dynamics into a network. The question of how well it must match the robot then becomes a question about the network, which is what we test.

\textbf{Compounding error.} Errors that grow along a rollout are a known weakness of learned dynamics models~\cite{r6,r7,r9}. The established remedies are to train the model on its own multi-step predictions~\cite{r6,r7,r8,r10} and to keep rollouts short~\cite{r5}. \emph{Teacher forcing} is the opposite practice: every step is predicted from the true previous state~\cite{r12}. One-step accuracy need not track how well a model serves control, a point known as objective mismatch~\cite{r11}. So the direction of RWM's training claim is expected. What is open is whether it holds on held-out data at this scale, and what a one-step check would have said.

\textbf{Uncertainty for model-based control.} Ensembles are the standard way to estimate what a model does not know~\cite{r17,r18}. Offline methods penalise imagined reward by an ensemble estimate~\cite{r13,r14,r15}, or pick models by their error on held-out single transitions~\cite{r16}. Lu et al.~compare such penalties by how well they track true error and separate ranking from scale, which is the closest precedent for Section 6~\cite{r19}. Recalibration corrects predicted uncertainty on held-out data~\cite{r20,r21} and resembles split conformal prediction~\cite{r22}. Variance heads trained by likelihood have known failure modes~\cite{r23}; the released loss is not a likelihood at all. Ensembles whose heads share one trunk are used in reinforcement learning~\cite{r24}, and such heads are known to under-report disagreement~\cite{r53,r54}.

\textbf{Stability of recurrent models.} Products of Jacobians govern how perturbations grow in a recurrent network~\cite{r27}. Contracting recurrent models are easier to train~\cite{r28}, Lyapunov spectra describe chaotic ones~\cite{r29}, and modified teacher forcing or built-in stability address the training difficulty~\cite{r30,r31,r32}. We use the leading finite-time Lyapunov exponent, computed by Benettin's method~\cite{r26}, as a plain measurement of whether a simulator amplifies its own errors.

\textbf{Physical consistency.} A predicted state can be checked against relations it must satisfy, as in classical fault detection~\cite{r34}. Schäfer et al.~ask whether imagined positions follow from imagined velocities, and diagnose one model on one task~\cite{r33}. We use a similar residual to rank many simulators and to test alarms and repairs.

\textbf{Compression.} Pruning and quantisation are standard ways to shrink a network~\cite{r35,r37}, and a small network trained from scratch often matches a pruned one~\cite{r36}. Recent work quantises world models and judges them by task success or by one fitted error-growth rate~\cite{r38,r39,r51}. None reports the error added at each horizon of a rollout, prunes, or measures cost on a CPU. Learned dynamics models already run inside embedded model-predictive control~\cite{r50}.

\textbf{How results are reported.} With few seeds and few trajectories, intervals must resample the unit that is actually independent~\cite{r40,r41,r42}. We register decision rules before the data exist~\cite{r43}, and report the count of independent units beside every interval.

\textbf{Our earlier report.} The rebuild and the uncertainty study appeared as a preprint~\cite{r3}. This paper adds the comparison of pre-deployment checks, the compression study and the robot-level tests, and puts them together for the first time.

\section{Robot, data and how we measure}

\subsection{The robot and its state}

ANYmal D is a quadruped with three actuated joints per leg: hip abduction, hip flexion and knee, 12 in all. The simulator works on the robot's own body measurements, with no camera or terrain input. Its state has 45 numbers: the body's linear velocity (3) and angular velocity (3), the direction of gravity seen from the body (3), and each joint's angle, speed and torque (12 each). The 12 joint commands are sent 50 times a second. A \emph{velocity command} is the walking speed and turning rate the robot is asked to follow.

\subsection{The released recording}

The release contains one file of 10,000 rows, which is 200 s of walking. We analysed it before training anything.

\begin{itemize}
\tightlist
\item
  \textbf{It is ten episodes of 20 s, and nothing in the file says so.} At ten rows, every joint speed and every joint command is exactly zero. These are resets. The column meant to mark the end of an episode is zero everywhere.
\item
  \textbf{The data are simulated.} An exact reset cannot come from hardware.
\item
  \textbf{The robot never falls.} No row is flagged as a fall, and a thigh touches the ground in one row out of 10,000.
\item
  \textbf{One gait.} The robot trots throughout, with a stride of 27 steps (0.54 s, 1.85 Hz).
\item
  \textbf{About twenty velocity commands.} The command changes once in the middle of nearly every episode, giving 21 segments of roughly 10 s. A held-out episode therefore tests other velocity commands of the same gait, and nothing more.
\item
  \textbf{Commands lead motion.} The joint command leads the joint angle by 4 steps (80 ms), as a position controller with a delay would produce.
\end{itemize}

The file is small. The eight training episodes of our main split hold 7,991 transitions, which is 0.13\% of the 6 million that RWM reports for pretraining.

\subsection{The simulator}

The network is RWM's. Two layers of gated recurrent units (GRU, a standard recurrent layer) of width 256 read the last 32 states and joint commands. A mean head predicts the change to the next state, and a second head predicts a variance. A second recurrent network predicts foot contacts and the end of the episode. The single-network model has 1,417,789 parameters. The released checkpoint has five output heads on one shared trunk and 1,995,569 parameters; the spread between its heads is the uncertainty RWM-U uses.

Unless stated otherwise, our models were trained with the release's loss and the Adam optimiser~\cite{r25} (learning rate 1e-4, weight decay 1e-5, batches of 256 windows) for 2,500 iterations, the number the paper gives.

\subsection{Pools of simulators and independent units}

Two 400-step trajectories that overlap are not independent evidence. Each held-out episode yields exactly two non-overlapping trajectories of 32 history steps and 368 forecast steps. That count, and not the number of windows, is the sample size of every long-horizon claim.

\begin{center}\small
\begin{tabularx}{\textwidth}{@{}>{\raggedright\arraybackslash\hsize=0.744\hsize}X>{\raggedright\arraybackslash\hsize=1.209\hsize}X>{\raggedright\arraybackslash\hsize=0.837\hsize}X>{\raggedright\arraybackslash\hsize=1.209\hsize}X@{}}
\toprule
\textbf{Pool} & \textbf{Held-out data} & \textbf{Independent trajectories} & \textbf{Simulators} \\
\midrule
Main pool & episodes 1 and 8 & 4 & 97: 57 trained, 40 compressed copies \\\addlinespace[2pt]
Five-split pool & every episode held out once, in five fixed pairs & 20 & 255: 55 trained, 200 compressed copies \\\addlinespace[2pt]
Other architectures & episodes 1 and 8 & 4 & 38: 13 MLP, 12 RSSM, 13 Transformer \\\addlinespace[2pt]
Released checkpoint & none: its training rows are not recorded & 20, never called held out & 1, and 8 compressed copies \\\addlinespace[2pt]
MuJoCo test sets & held-out MuJoCo episodes & 40 expert windows, 67 mixed & 4 trained from scratch; 18 fine-tuned copies of the released model \\
\bottomrule
\end{tabularx}
\end{center}

Eleven trained simulators are in both of the first two pools, so the project counts 101 trained recurrent simulators. The main pool contains five seeds of the released configuration, ensembles, a likelihood loss, other history and forecast lengths, longer training, and nine models trained one step ahead. Compressed copies are made after training: weights rounded to 16 or 8 bits, the smallest 10 to 50\% of weights set to zero (\emph{pruning}), or recurrent units removed.

\subsection{Measures}

\textbf{Prediction error.} $E(h)$ is the mean absolute error over the 45 state numbers, each divided by its standard deviation in the data, averaged over rollouts of $h$ steps under the recorded joint commands. We write $E_{1}$, $E_{8}$, $E_{100}$ and $E_{368}$ for its values at 1, 8, 100 and 368 steps. The rebuild study~\cite{r3} also reports the release's own \emph{relative L1} error, and Sections 4 and 6 use it where we compare with the release. We name the measure wherever it changes.

\textbf{The floor.} The simplest forecast predicts no change and holds the last state. A simulator that does worse than this at some horizon has no skill there. The \emph{skill retained} by a compressed copy, $K(h)$, is the share of its parent's advantage over the floor that it keeps: \[
K(h) = 1 - \frac{\Delta(h)}{S(h)}
\] Here $\Delta(h)$ is the copy's extra error over its parent, and $S(h)$ is the parent's advantage over the floor. A negative $K$ means worse than predicting no change.

\textbf{Two checks that need no recorded future.} The \emph{growth rate} $\lambda_1$ is the leading finite-time Lyapunov exponent. Take two copies of the simulator that start from almost identical internal states and receive the same joint commands. $\lambda_1$ is the average rate per second at which the gap between them grows (positive) or shrinks (negative, a \emph{contracting} simulator). The \emph{kinematic residual} $\bar{r}_J$ asks whether the predicted joint angles change as the predicted joint speeds say they should, in units of the same residual in the recorded data.

\textbf{Uncertainty.} \emph{Coverage at one sigma} is the share of outcomes whose error is within the predicted standard deviation; a calibrated Gaussian estimate gives 68\%. The \emph{error-to-spread ratio} is the mean absolute error over the mean predicted standard deviation; a calibrated estimate gives 0.80.

\textbf{Robot-level outcomes.} A policy is run in the physics model. The \emph{fall rate} is the share of episodes that end in a fall. The \emph{physics return} is the training code's own reward, computed from physics states. The \emph{walking score} gives each step the value \[
\exp\!\left(-\frac{\lVert v_{xy}-c_{xy}\rVert^2 + (\omega_z - c_\omega)^2}{0.25}\right)
\] where $v$ and $\omega$ are the body's speed and turning rate, and $c$ is the velocity command, with zero for every step after a fall. It is averaged over 96 episodes of 20 s. The authors' released policy scores 0.942 and standing still scores 0.069, so the score separates walking from merely staying upright.

\subsection{Statistics and registration}

Intervals are 95\% percentile intervals from 20,000 bootstrap draws over the independent unit: whole trajectories when two methods are compared, and models when checks are correlated with error across simulators~\cite{r40,r41}. Primary questions use 97.5\% intervals. With four trajectories, an exact sign-flip test cannot give a two-sided $p$ below 1/8, so no comparison at that size is called resolved. Holm's procedure controls multiple tests within a registered family, not across studies.

A result is \emph{registered} if its decision rule was committed before the measurements that test it were run~\cite{r43}. A rule gives a verdict only when its interval clears a stated threshold, and `not settled' otherwise. The rebuild study alone committed 35 rules~\cite{r3}; each later study registered its own. The registrations are commits in our own version control, so their times are self-reported.

\subsection{Compute}

Training and all policy work ran on an Intel Core i5-7360U laptop from 2017, with two cores and 8 GiB of memory, using PyTorch 2.2.2. A single-network model trains at about 1.3 s per iteration, so 2,500 iterations take under an hour; the five-head ensemble takes about 4.5 hours. One policy takes 1.8 to 2.7 hours. The analysis-only studies of Section 7 were computed in a two-core Linux workspace and re-checked on the laptop, where last digits can differ.

\section{Rebuilding the model and auditing the release}

Our rebuild equals the released module exactly, so every later difference belongs to the method or the data, and not to our wiring. The audit that made this possible also found two defects in the released pipeline and several places where the code departs from the papers.

\subsection{Verification before any training}

Each level was passed before the next was attempted~\cite{r3}.

\begin{center}\small
\begin{tabularx}{\textwidth}{@{}l>{\raggedright\arraybackslash\hsize=1.000\hsize}X>{\raggedright\arraybackslash\hsize=1.000\hsize}X@{}}
\toprule
\textbf{Level} & \textbf{What was checked} & \textbf{Result} \\
\midrule
Shapes & parameter counts against the released checkpoint & exact \\\addlinespace[2pt]
Wiring & outputs against the released module & difference 0, bit for bit \\\addlinespace[2pt]
Indexing & the test harness feeds the joint commands it claims & bit for bit against the raw file \\\addlinespace[2pt]
Residual & a model that predicts zero change equals the floor & 1.19e-07 \\\addlinespace[2pt]
Objective & losses and gradients against the released code & difference 0 across 7 loss terms and 106 parameter tensors \\\addlinespace[2pt]
Trainer & can memorise a single batch & loss falls 1,506-fold \\
\bottomrule
\end{tabularx}
\end{center}

The objective check is the important one. Reading the configuration file would not have found what follows; reading the source line by line and comparing gradients did.

\subsection{Where the code departs from the papers}

None of these is hidden, but none is stated in the papers, and each matters to someone who wants to reuse the model~\cite{r3}.

\begin{itemize}
\tightlist
\item
  \textbf{Seven loss terms, four active.} The paper describes two. Of the seven in the code, three are switched off by the released settings.
\item
  \textbf{The contact and fall predictor never sees a rollout.} The state network is trained on its own 8-step predictions. The second network, which predicts foot contacts and the end of the episode, is trained on true states only. In imagination, a fall is whatever that predictor says it is.
\item
  \textbf{No decay factor.} The paper weights later forecast steps by a decay factor; the code has none.
\item
  \textbf{Three iteration counts.} The configuration says 500, the paper 2,500, and the released checkpoint is tagged 5,000.
\item
  \textbf{A standard deviation where the paper writes a variance.} RWM-U's penalty is written as a variance across the ensemble; the code computes a standard deviation. The first author confirmed that the code is what was intended~\cite{r3}.
\item
  \textbf{Only one of two uncertainties is used.} The code computes each network's own predicted variance and the spread between networks. Only the spread reaches the policy; the other is computed and never read.
\end{itemize}

\subsection{Two defects in the released pipeline}

\textbf{Episode boundaries are not marked.} The data file joins ten episodes, and its end-of-episode column is always zero. The release's window builder therefore accepts all 9,961 windows of 40 rows, of which 352 (3.5\%) glue the end of one episode onto the start of the next. We recover the boundaries from the reset rows and keep 9,609 windows. Training on the glued windows cost nothing we could measure: over 32 comparisons, a contaminated model was hurt in none and helped in 9, which fits mild regularisation. It is still a defect, because glued windows can carry held-out rows into training.

\textbf{Evaluation feeds each command one step late.} Row $t$ of the file holds the joint command that produced state t. We know this because at every reset row all twelve commands are exactly zero, which a policy network with biases cannot output. Training pairs each state with the command in its own row; the released evaluation pairs it with the command from the row before. On the released checkpoint, this raises the release's relative L1 error by 55\% {[}26\%, 93\%{]} one step ahead and by 37\% {[}21\%, 65\%{]} at 32 steps. From 100 steps on, the change is not resolved (8.2\% {[}−10\%, 25\%{]}), on 20 trajectories. We reported it to the first author.

The second defect is a robotics lesson in its own right. An off-by-one in when a command takes effect is invisible in the code and in aggregate plots. It was found by asking what the recorded numbers could physically mean.

\subsection{What cannot be known about the released checkpoint}

\textbf{Its training data are not recorded.} No released file says which rows trained the checkpoint. Its one-step error is the same on the eight episodes our models trained on and on the two they did not (0.083 against 0.085), whereas our own model's differs (0.075 against 0.12). So we never call any measurement of the released checkpoint held out.

\textbf{Its variance state cannot be reached from the released starting point.} Section 6.2 shows that the predicted variance shrinks at a steady rate during training. Even at the largest step the optimiser can take, reaching the checkpoint's stored value would need 45,737 iterations, against the 2,500 in the paper. The first author's account is that the release is several revisions removed from the setup that trained the checkpoint. We read this as a gap in documentation, not as an inconsistency~\cite{r3}.

\section{Training on rollouts against training one step ahead}

RWM's central training claim holds on data the models never saw: at 368 steps, training on the model's own rollouts beat teacher forcing on all ten held-out episodes. One step ahead, the order reverses. That reversal is the reason one-step checks mislead, and it has a mechanical explanation.

\subsection{The registered result}

We trained both methods on five fixed splits of the ten episodes, so that each episode is held out exactly once (2,500 iterations, three seeds per method per split). This gives 20 independent held-out trajectories.

\begin{itemize}
\tightlist
\item
  \textbf{At 368 steps, rollout training wins everywhere.} Teacher forcing's relative L1 error exceeds rollout training's by +2.2 {[}+1.4, +3.5{]}, a factor of 5.9. All 10 of 10 episodes favour rollout training (sign-flip test over episodes, $p = 0.002$; Figure 2). The verdict of a rule committed before these runs was `replicates across folds'.
\item
  \textbf{One step ahead, teacher forcing wins, on short windows.} On 300 non-overlapping 33-row units, the difference is −0.0090 {[}−0.013, −0.0053{]} in favour of teacher forcing. Episode by episode, the sign agrees on only 6 of 10, so this is a short-window result.
\item
  \textbf{Longer training does not change the picture.} At 10,000 iterations, on the main held-out pair, the factor at 368 steps is 4.6 over three seeds, and 2.58 at 100 steps.
\item
  \textbf{Teacher forcing ends up worse than predicting nothing.} At 10,000 iterations and 368 steps, the teacher-forced model is 1.66 times worse than holding the last state, while the rollout-trained model is 2.8 times better~\cite{r3}.
\end{itemize}

\begin{figure}[tbp]
\centering
\includegraphics[width=0.50\linewidth]{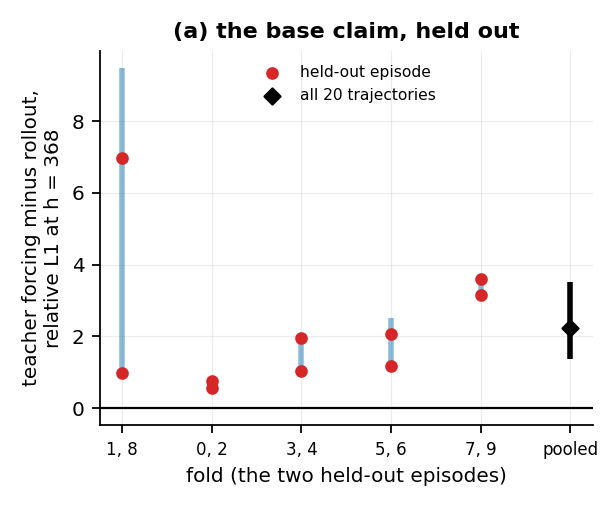}
\caption{Teacher forcing's error minus rollout training's at 368 steps (relative L1), for each of the five splits. Each red dot is one held-out episode, and every dot is above zero. The black diamond is the difference pooled over all 20 trajectories, with its 95\% interval.}\label{fig:2}
\end{figure}

\subsection{How this claim was measured: three attempts}

The verdict above is the third one this project reached, and the first two are part of the result.

\begin{enumerate}
\def\labelenumi{\arabic{enumi}.}
\tightlist
\item
  \textbf{Asserted at 8 steps.} Our first rule tested the claim at the horizon the model is trained on. It passed, using seed spread and overlapping trajectories.
\item
  \textbf{Withdrawn.} Measured again on independent trajectories only, the same rule returned `cannot be settled' in all eight combinations of data, trajectory length and metric. The rule had been tied to the training horizon, where the two methods have barely separated, and not to the long horizon the claim is about.
\item
  \textbf{Re-tested at 368 steps under a new rule committed before the runs.} All three conditions were met: rollout training led at both checkpoints, the interval on the gap excluded zero ({[}+0.56, +2.05{]}, four trajectories), and the sign was positive on all ten episodes.
\end{enumerate}

The five-split result of Section 5.1 then replicated it with twenty trajectories.

\subsection{What a one-step check would have said}

The table gives the same comparison in the normalised error $E(h)$, on the five-split pool, at the horizons a user cares about: 24 steps is 0.48 s, 100 steps is the 2 s that RWM imagines, and 368 steps is the release's test length. A ratio above 1 means the one-step choice is worse (descriptive).

\begin{center}\small
\begin{tabularx}{\textwidth}{@{}>{\raggedright\arraybackslash\hsize=1.000\hsize}Xrrrr@{}}
\toprule
\textbf{Choice} & \textbf{1 step} & \textbf{24 steps} & \textbf{100 steps} & \textbf{368 steps} \\
\midrule
Error of teacher forcing over rollout training & 0.90 & 1.5 & 2.0 & 5.0 \\
Error of training horizon 1 over training horizon 8 & 0.86 & 1.8 & 2.6 & 5.4 \\
\bottomrule
\end{tabularx}
\end{center}

One step ahead, both one-step choices look better: their ratios are below 1. By 8 steps, teacher forcing is worse on 19 of the 20 trajectories, and it crosses the floor at step 29 (Figure 3).

\begin{figure}[tbp]
\centering
\includegraphics[width=0.68\linewidth]{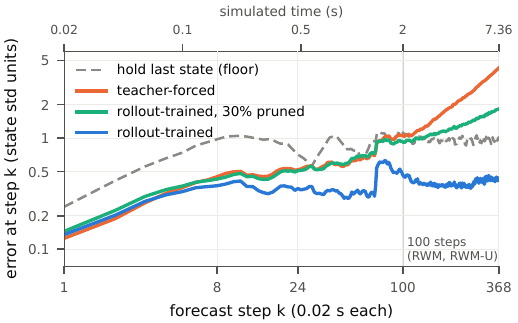}
\caption{Error at each forecast step on the 20 held-out trajectories of the five splits, with models averaged within each trajectory. The teacher-forced model first exceeds the floor at step 29, and the 30\%-pruned copy at step 57. The vertical line marks the 100-step horizon that RWM and RWM-U imagine.}\label{fig:3}
\end{figure}

In physical units, the gap is plain. Over 2 s, the teacher-forced simulator's joint angles are off by 7.0 degrees against 3.4 degrees, and it misplaces the robot by 0.84 m against 0.51 m, out of 1.5 m travelled. A caution belongs here too: over 2 s, neither simulator beats the assumption that the robot keeps its last measured velocity (0.45 m). By 7.4 s, rollout training beats it and teacher forcing does not.

\subsection{Why: the training horizon sets whether errors grow}

The growth rate $\lambda_1$ measures whether a simulator amplifies a small error in its own state. It follows the training horizon closely (registered).

\begin{center}\small
\begin{tabular}{@{}lrrrrr@{}}
\toprule
\textbf{Steps predicted per training window} & \textbf{1} & \textbf{2} & \textbf{8} & \textbf{16} & \textbf{32} \\
\midrule
Growth rate $\lambda_1$ (per second) & +0.61 & +0.44 & −0.43 & −1.13 & −1.38 \\
\bottomrule
\end{tabular}
\end{center}

\begin{figure}[tbp]
\centering
\includegraphics[width=0.62\linewidth]{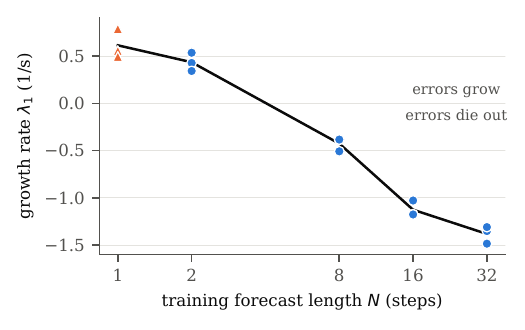}
\caption{Growth rate against the training horizon, for a history of 32 steps and three seeds each. Triangles mark models trained one step ahead. Above zero, a small error grows; below zero, it dies out.}\label{fig:4}
\end{figure}

The rank correlation between training horizon and growth rate is −0.97 {[}−0.99, −0.89{]} over 15 models. Models trained one or two steps ahead amplify their own errors; models trained on eight or more of their own steps damp them (Figure 4). Teacher forcing shifts the growth rate by +1.18 per second {[}1.07, 1.30{]} relative to rollout training (registered). Every rebuilt model trained on rollouts of eight or more steps was contracting, and no teacher-forced one was.

This explains the reversal. A one-step check gives the simulator true recorded states. In use, it reads its own predictions, so what matters is how it treats its own errors. Teacher forcing never shows the network those errors.

The released checkpoint does not fit this account. Its growth rate is +0.70 per second on these trajectories, yet its errors are low. It was trained on data we do not have, about 750 times as much as ours, and may have seen these very trajectories.

\subsection{Other networks, longer training and the weaker claims}

\begin{itemize}
\tightlist
\item
  \textbf{Other architectures} (descriptive). The reversal held in 5 of 5 seed pairs for a multilayer perceptron, and in 5 of 5 for a Transformer. For a recurrent state-space model (RSSM), teacher forcing was better one step ahead in 5 of 5 pairs, but worse at 100 steps in none.
\item
  \textbf{Four times longer training} (descriptive, three seeds). The 100-step error fell for both methods: from 0.251 to 0.181 with rollout training, and from 0.576 to 0.464 with teacher forcing. In every seed, teacher forcing stayed better one step ahead and worse at 100 steps.
\item
  \textbf{The configuration claim.} RWM presents a history of 32 and a forecast of 8 as a good trade-off. At our budget, the centre was not shown best and no alternative was shown better: four trajectories cannot separate them. A forecast of 32 had the highest one-step error (0.10) and the lowest 100-step error (0.19).
\item
  \textbf{The architecture claim.} RWM says its model beats an MLP, an RSSM and a Transformer. Every baseline we built was worse than our rollout-trained model on all four trajectories at 368 steps. That is consistent with the claim and a weak test of it: the baselines are untuned, and none beat even the plain forecast of the training mean~\cite{r3}.
\end{itemize}

\section{What the uncertainty reports}

RWM-U's uncertainty ranks rollout errors well and sizes them wrongly. On the released checkpoint at 100 steps, the disagreement between its five heads has a rank correlation of +0.90 with the error, yet covers 4.6\% of outcomes at one sigma, where a calibrated estimate covers 68\%. A policy that reads it as a distance to the truth is misled; a policy that reads it as an ordering is not.

\subsection{Two uncertainties, one of them used}

At every step, the code computes two numbers. The first is the variance each head predicts for its own output, meant to capture noise in the robot itself. The second is the spread between the five heads' predictions, meant to capture what the model does not know. Only the second reaches the policy, as a penalty subtracted from the reward~\cite{r2,r13}. We measured both.

\subsection{Each head's own variance collapses, by construction}

The loss that trains the variance is not a likelihood. It is the squared error of a random sample drawn from the predicted distribution. Its expected value is \[
\mathbb{E}\big[(\mu + \sigma\varepsilon - y)^2\big] = (\mu - y)^2 + \sigma^2
\] which is smallest when the predicted standard deviation $\sigma$ is zero, whatever the data. Nothing else in the loss pushes back. We checked this in three ways.

\begin{itemize}
\tightlist
\item
  \textbf{On data with known noise.} With synthetic noise that varies 25-fold across the input range, the released head under the released loss sits 22 times below the true noise and does not follow it. Under the authors' own unused likelihood option, it recovers the right level (ratio 0.98).
\item
  \textbf{In the speed of the collapse.} Across 81 training runs, the variance bound shrinks by a median of 0.94 learning rates per iteration, which is simply the optimiser's ordinary step. Under the likelihood option the sign flips (Figure 5).
\item
  \textbf{In its size.} On the released checkpoint, each head's own standard deviation is 1,827 times smaller than its error one step ahead, and 11,683 times smaller at 100 steps.
\end{itemize}

\begin{figure}[tbp]
\centering
\includegraphics[width=0.96\linewidth]{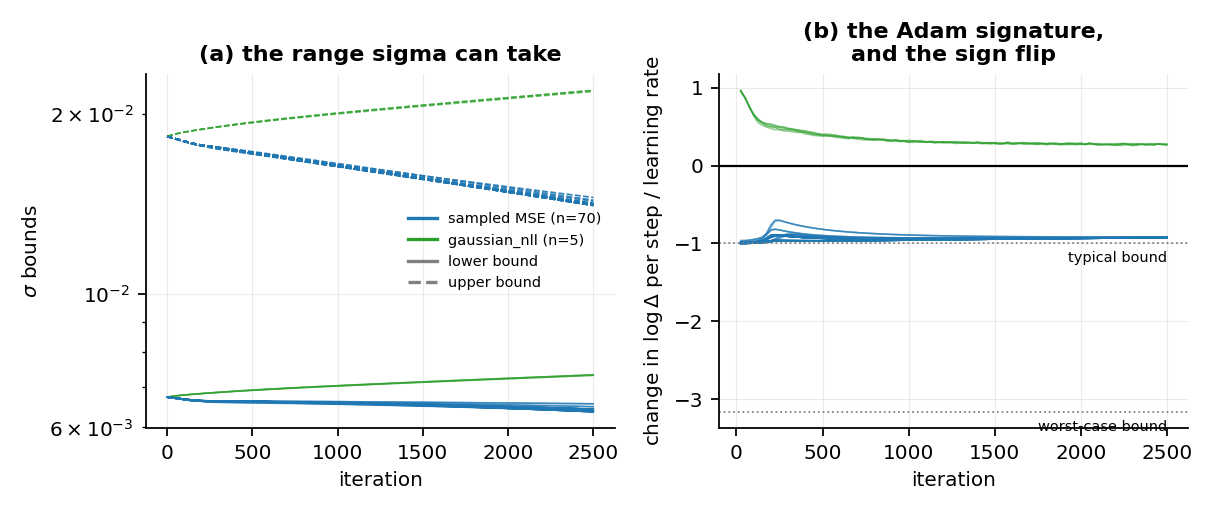}
\caption{The collapse during training, in 75 logged runs of 2,500 iterations: 70 with the released loss (blue) and 5 with the likelihood loss (green). (a) The lowest (solid) and highest (dashed) standard deviation the head can output. Under the released loss both fall; under the likelihood loss both rise. (b) How fast the gap between the two bounds changes, in learning rates per iteration, beside the optimiser's typical and largest step. The likelihood loss flips the sign.}\label{fig:5}
\end{figure}

RWM-U reads the small value as a sign that the robot has little noise. The observation is right and the reading is not. Switching to the likelihood loss reverses the collapse, but does not give a usable estimate (Figure 6): the predicted value stays almost the same for every input~\cite{r3}. The bounded variance head comes from an earlier method~\cite{r17}. Of 13 public code bases we inspected that descend from it, 10 implement the head and 1 can train it this way.

\begin{figure}[tbp]
\centering
\includegraphics[width=1.00\linewidth]{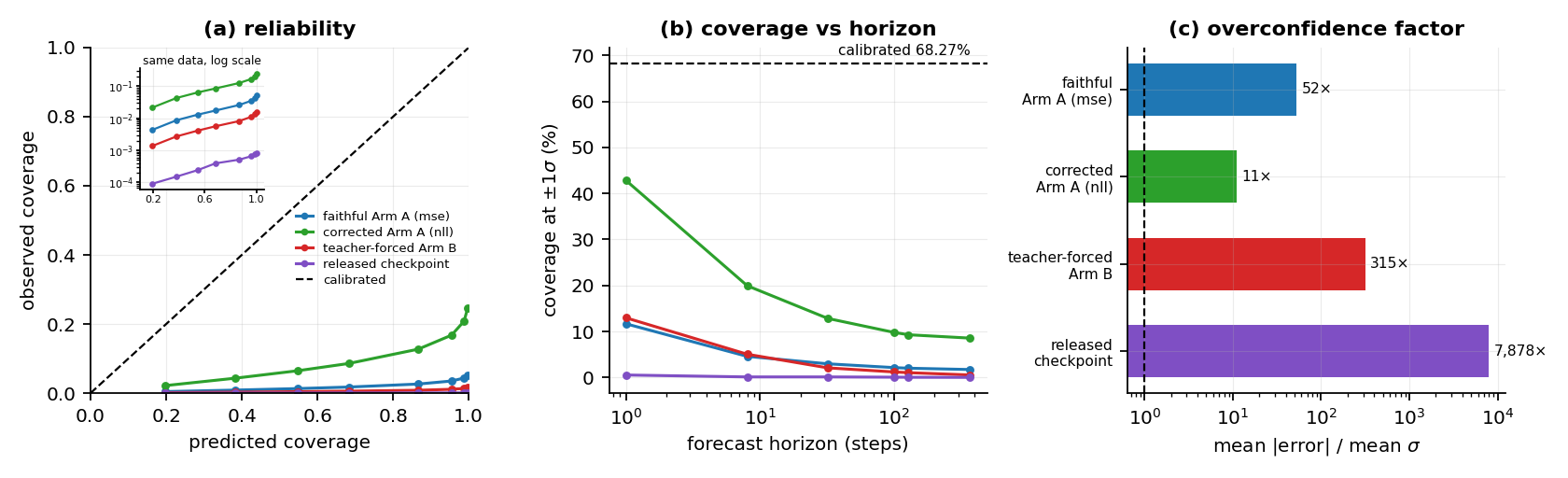}
\caption{Calibration of each network's own predicted standard deviation, for four models: our rollout-trained model with the released loss (`faithful Arm A'), the same with the likelihood loss (`corrected Arm A'), our teacher-forced model (`Arm B') and the released checkpoint. The data are the four held-out trajectories of the main split, which may be in-sample for the released checkpoint. (a) Observed against predicted coverage; a calibrated estimate lies on the diagonal. (b) Coverage at one sigma against forecast horizon; a calibrated estimate gives 68\%. (c) Mean absolute error over mean predicted standard deviation; the dashed line is where the two are equal. Every model is far too confident.}\label{fig:6}
\end{figure}

\subsection{The spread between heads: right ranking, wrong scale}

The table gives the released checkpoint on all ten released episodes, 20 independent trajectories. It may have been trained on them, so these are not held-out numbers.

\begin{center}\small
\begin{tabularx}{\textwidth}{@{}>{\raggedright\arraybackslash\hsize=1.000\hsize}Xrrrrr@{}}
\toprule
\textbf{Forecast horizon (steps)} & \textbf{1} & \textbf{8} & \textbf{32} & \textbf{100} & \textbf{368} \\
\midrule
Mean error over mean spread (calibrated: 0.80) & 8.3 & 15.1 & 22.6 & 33.4 & 34.4 \\
Coverage at one sigma (calibrated: 68\%) & 16.2\% & 10.0\% & 7.0\% & 4.6\% & 3.6\% \\
Coverage at two sigma (calibrated: 95\%) & 30.1\% & 19.8\% & 13.7\% & 9.3\% & 7.2\% \\
\bottomrule
\end{tabularx}
\end{center}

\textbf{As a ranking, it holds up.} The spread ranks error better than the forecast step number does, at every horizon from 8 steps on. That comparison matters because the step number is free: errors grow along any rollout, so any signal that grows with time will correlate with error. With the rollout and the forecast depth both held fixed, the spread still correlates +0.42 {[}+0.32, +0.58{]} with the error. Against the model's own predicted step size, another free signal, a registered test on 70 shorter units at 100 steps finds the spread ahead by +0.28 {[}+0.20, +0.39{]}.

\textbf{As a scale, it fails at every horizon, and the failure grows.} The factor by which the spread must be multiplied to cover 68\% of outcomes is 6.6 one step ahead, 31 at 100 steps and 42 at 368 steps. A one-step calibration check therefore understates the problem at the horizon of use.

\subsection{Three separate reasons the scale is wrong}

\begin{enumerate}
\def\labelenumi{\arabic{enumi}.}
\tightlist
\item
  \textbf{The spread falls short even one step ahead.} At one step nothing has accumulated, yet the squared error exceeds what the spread implies by a factor of 174 (95\% interval 82 to 255) on the released checkpoint, and 7.6 {[}4.0, 9.7{]} on our own shared-trunk ensembles.
\item
  \textbf{The five heads share one trunk.} Per member, 89\% of the parameters are identical to every other member's, and one recurrent state serves them all. They can disagree only about what that shared state already carries. Five independently trained models, used as an ensemble, were 2.0 times better calibrated at 100 steps, against a smallest detectable gain of 1.5 fixed in advance. They were still far too confident.
\item
  \textbf{The release feeds every head the ensemble mean.} The spread at a deep step is then a one-step spread at one shared history, so it cannot track accumulated error. Rolling each head forward on its own predictions closes 0.75 {[}0.69, 0.82{]} of the gap on a log scale, and raises coverage at 100 steps from 4.6\% to 33\% (Figure 7). This is not the released method.
\end{enumerate}

\begin{figure}[tbp]
\centering
\includegraphics[width=0.50\linewidth]{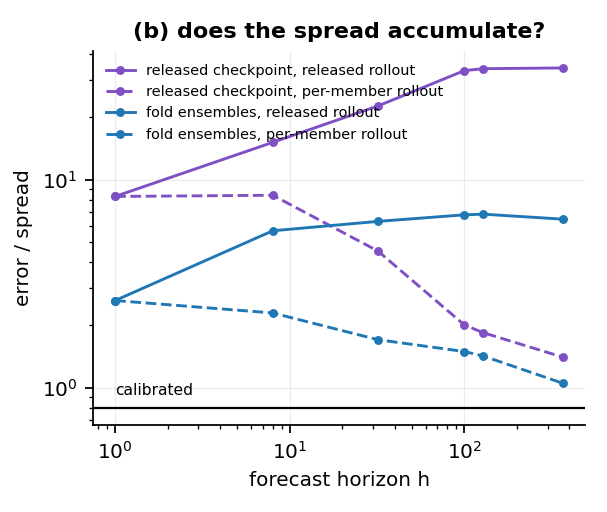}
\caption{Error over spread against forecast horizon, for the released checkpoint and for our five-split ensembles (20 trajectories each). Solid lines use the released evaluation's rollout, which feeds the ensemble mean back in. Dashed lines roll each member out on its own predictions, which the release does not do. A calibrated estimate lies on the line at 0.80.}\label{fig:7}
\end{figure}

\subsection{Can it be repaired after training?}

Partly. One multiplier per horizon, fitted on nine episodes and scored on the tenth, brought the released checkpoint's one-sigma coverage within ten points of the target at all six horizons (registered). It did not repair the tails: two-sigma coverage at 100 steps was 84\% {[}81\%, 86\%{]} against 95\%, because the errors are heavy-tailed. On episodes a model never saw, the evidence is mixed: cell by cell, the multiplier often missed, although pooled coverage landed near the target~\cite{r3}.

Compression does not disturb the ranking. Rounding the weights to 8 bits changed the rank correlation between spread and error by −0.0065 at 8 steps and −0.0039 at 100 steps (exploratory).

\subsection{What this means for a reward penalty}

A penalty weight can absorb any constant factor, so a spread that is too small by a fixed amount would do no harm. A factor that grows along the rollout, from 6.6 to 31 by the horizon of use, cannot be absorbed by one weight. Section 9 shows the practical side: in simulators whose heads disagree more than the released model's, the same penalty drove policies to end their imagined episodes at once.

\section{Which check should be run before a policy is trained?}

Among checks of prediction error, judge a learned simulator by its error eight or more steps ahead on held-out episodes. One-step error ranked our main pool of simulators poorly, and ranked its trained models backwards. Two checks that need no recorded future helped in places, but neither was shown better than 8-step error in a primary test, and neither worked as an alarm during a rollout. None of these checks is shown here to predict whether a policy trained in the simulator walks: on the fine-tuned simulators of Section 9.5, error one and eight steps ahead ranked them partly in reverse.

\subsection{Four checks, one target}

A check that needs no recorded future would be valuable to a roboticist. It could vet a simulator on new velocity commands, or while a policy drives it, where no recording exists. We compared four checks by how well they rank simulators by their 100-step error under recorded commands:

\begin{itemize}
\tightlist
\item
  one-step error $E_{1}$ and 8-step error $E_{8}$, which need recorded future states;
\item
  the growth rate $\lambda_1$ and the kinematic residual $\bar{r}_J$, which do not.
\end{itemize}

Rankings use Spearman's rank correlation across simulators. The comparisons were registered before their measurements were run.

\subsection{How each check ranks simulators}

Correlation of each check with the 100-step error (descriptive; number of simulators in brackets).

\begin{center}\footnotesize
\begin{tabularx}{\textwidth}{@{}>{\raggedright\arraybackslash\hsize=1.250\hsize}X>{\centering\arraybackslash\hsize=1.150\hsize}X>{\centering\arraybackslash\hsize=1.150\hsize}X>{\centering\arraybackslash\hsize=1.000\hsize}X>{\centering\arraybackslash\hsize=0.750\hsize}X>{\centering\arraybackslash\hsize=0.750\hsize}X>{\centering\arraybackslash\hsize=0.950\hsize}X@{}}
\toprule
\textbf{Check} & \textbf{Main pool, recorded commands (97)} & \textbf{Main pool, policy's commands (97)} & \textbf{Five-split pool (255)} & \textbf{MLP (13)} & \textbf{RSSM (12)} & \textbf{Transformer (13)} \\
\midrule
One-step error $E_{1}$ & +0.32 & not applicable & +0.78 & −0.69 & +0.88 & −0.61 \\
8-step error $E_{8}$ & +0.84 & not applicable & +0.95 & +0.76 & +0.86 & +0.73 \\
Growth rate $\lambda_1$ & +0.80 & +0.76 & +0.67 & +0.84 & +0.26 & +0.87 \\
Kinematic residual $\bar{r}_J$ & +0.95 & +0.83 & +0.96 (exploratory) & +0.76 & +0.94 & +0.98 \\
\bottomrule
\end{tabularx}
\end{center}

\begin{figure}[tbp]
\centering
\includegraphics[width=0.80\linewidth]{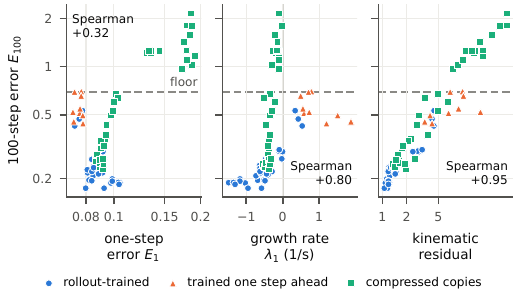}
\caption{One-step error, growth rate and kinematic residual (left to right) against 100-step error, for the 97 simulators of the main pool. Models trained one step ahead (triangles) have low one-step error, high 100-step error and positive growth rates. Compressed copies (squares) spread mostly up and down. The dashed line is the floor.}\label{fig:8}
\end{figure}

\subsection{Registered verdicts}

Each row is the difference between two rank correlations with the 100-step error, first check minus second. Primary questions (P) use 97.5\% intervals, and secondary ones (S) 95\%.

\begin{center}\small
\begin{tabularx}{\textwidth}{@{}>{\raggedright\arraybackslash\hsize=1.350\hsize}X>{\raggedright\arraybackslash\hsize=0.650\hsize}X>{\raggedright\arraybackslash\hsize=1.000\hsize}X>{\raggedright\arraybackslash\hsize=1.000\hsize}X@{}}
\toprule
\textbf{Question} & \textbf{Pool} & \textbf{Estimate} & \textbf{Verdict} \\
\midrule
Q1 (P): growth rate against $E_{1}$ & main & +0.48 {[}+0.20, +0.79{]} & growth rate better \\\addlinespace[2pt]
Q2 (P): growth rate against $E_{8}$ & main & −0.04 {[}−0.20, +0.14{]} & not settled \\\addlinespace[2pt]
Q3 (S): growth rate against $E_{1}$ & five-split & −0.12 {[}−0.22, −0.03{]} & $E_{1}$ better: replication failed \\\addlinespace[2pt]
Q3 (S): growth rate against $E_{8}$ & five-split & −0.28 {[}−0.37, −0.22{]} & $E_{8}$ better \\\addlinespace[2pt]
Q4 (S): residual against $E_{1}$ & main & +0.63 {[}+0.41, +0.88{]} & residual better \\\addlinespace[2pt]
Q4 (S): residual against $E_{8}$ & main & +0.12 {[}+0.02, +0.23{]} & residual better; fails multiple-test correction \\\addlinespace[2pt]
Q5 (S): Q1 without the one-step-trained models & main & +0.27 {[}+0.04, +0.51{]} & growth rate better; fails correction \\\addlinespace[2pt]
Q6 (S): does 30\% pruning raise the growth rate? & five-split & +0.03 {[}−0.02, +0.07{]} & not settled \\\addlinespace[2pt]
CL1 (P): residual, policy's commands minus recorded & main & −0.120 {[}−0.220, −0.052{]} & worse with policy commands \\\addlinespace[2pt]
CL2 (P): growth rate, policy's commands minus recorded & main & −0.032 {[}−0.113, +0.041{]} & not settled \\\addlinespace[2pt]
AR1 (P): growth rate against $E_{1}$ & other architectures & +0.796 {[}+0.368, +1.025{]} & growth rate better \\\addlinespace[2pt]
AR2 (P): residual against $E_{8}$ & other architectures & +0.110 {[}−0.081, +0.356{]} & not settled \\
\bottomrule
\end{tabularx}
\end{center}

\subsection{What the verdicts mean in practice}

\textbf{One-step error picks the wrong training choices.} Across the 57 trained models of the main pool, it ranks the 100-step error backwards (−0.50, exploratory), because the models trained one step ahead have the lowest one-step error and the highest long-horizon error (Figure 8). It ranks compressed copies of one model well (+0.89).

\textbf{The growth rate is a screen, not a certificate.} A positive growth rate flagged the simulators trained one step ahead. But the growth rate failed its replication on the five-split pool, where most simulators are compressed copies: moderate pruning adds error without a shift in growth rate that we could detect. A negative growth rate certifies nothing: of the 17 main-pool simulators that were worse than the floor at 100 steps, 15 had a growth rate at or below zero. It did not separate seeds of one configuration (0.16 {[}−0.10, 0.41{]}, exploratory). And the released checkpoint has a positive growth rate with low error.

\textbf{The kinematic residual ranked best where commands were recorded, and less well under a policy.} When the authors' released policy supplied the joint commands, reading the simulator's own predicted state, the residual's correlation fell from 0.95 to 0.83. The drop was carried by the trained models (−0.285 {[}−0.492, −0.137{]}). Small random noise on the recorded commands did not reproduce it.

\textbf{A candidate check that needs a policy but no recording.} Simulators with higher error gave the policy a lower imagined return (rank correlation −0.86 {[}−0.91, −0.77{]}, secondary). This was not registered as a check, and is a lead for follow-up.

\textbf{The intervals are optimistic.} Simulators in a pool share parents, data and test trajectories. Wider intervals that resample parents and trajectories together are 1.5 times as wide (median), and leave 3 of the 13 registered verdicts of the two recurrent pools unresolved (exploratory).

\subsection{Alarms and repairs during a rollout: negative results}

A check that ranks whole simulators need not flag individual bad steps. We asked whether any label-free signal can say, during a rollout, that the current step has become worse than holding the last state. None could.

\begin{center}\small
\begin{tabularx}{\textwidth}{@{}>{\raggedright\arraybackslash\hsize=0.850\hsize}X>{\raggedright\arraybackslash\hsize=1.400\hsize}X>{\raggedright\arraybackslash\hsize=0.750\hsize}X@{}}
\toprule
\textbf{Signal or intervention} & \textbf{Result} & \textbf{Reading} \\
\midrule
Combined kinematic score as an alarm & AUROC 0.54 {[}0.40, 0.68{]}; 0.5 is chance & no better than chance \\\addlinespace[2pt]
Growth rate read step by step & AUROC 0.50 against 0.64 for a plain step counter; difference −0.15 {[}−0.22, −0.03{]} & worse than counting steps (registered) \\\addlinespace[2pt]
Re-synchronising with the true state when the score fires & 10.1\% {[}5.6\%, 15.7\%{]} more 368-step error than the same number of re-synchronisations on a fixed schedule & a schedule is better (registered; Figure 9) \\\addlinespace[2pt]
Rescaling the predicted gravity direction to unit length & 368-step error changes by −3.5\% {[}−5.3\%, −2.1\%{]} & a small real gain \\\addlinespace[2pt]
Forcing joint angles to follow the integrated joint speeds & 368-step error multiplied by 2.3 & harmful \\
\bottomrule
\end{tabularx}
\end{center}

\begin{figure}[tbp]
\centering
\includegraphics[width=0.62\linewidth]{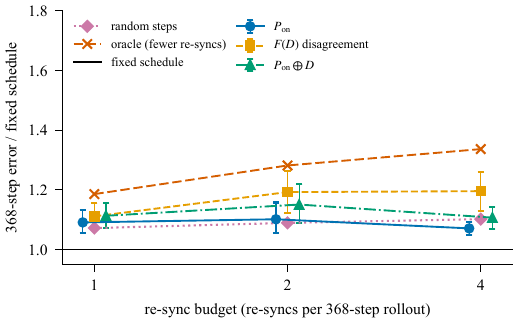}
\caption{Error at 368 steps relative to re-synchronising on a fixed schedule (the line at 1), against the number of re-synchronisations allowed per rollout. The triggers are the kinematic score computed during the rollout (circles), the calibrated ensemble disagreement (squares) and their average (triangles). Every point lies above the line: no trigger beat the schedule. Bars are 95\% intervals. The `oracle' re-synchronises where the plain rollout first fails; it uses fewer re-synchronisations, so it is not a like-for-like bound.}\label{fig:9}
\end{figure}

Two lessons follow. First, enforce a physical relation only where it is exact. The gravity direction has unit length by definition; the joint relation holds only approximately at 50 Hz, and enforcing it destroyed the forecast. Second, physical consistency is not accuracy. A simulator can produce a wrong state that is internally consistent, and a simulator that damps its errors can settle on a biased prediction.

\section{Making the simulator smaller and faster}

On our rebuilt models, rounding the weights to 8 bits was safe at every horizon. Removing 30\% of the weights without retraining looked safe one step ahead, and was worse than predicting no change by 368 steps. A short fine-tuning restored the lost skill. The released checkpoint tolerated the same compressions much less well, which is a warning against carrying such results from one model to another.

\subsection{Why compress, and how to judge it}

A simulator that runs on the robot, or on a small computer beside it, must be small and fast. Compression is normally judged by accuracy on single inputs. A rollout feeds every error back, so we judge each compressed copy by the skill it retains, $K(h)$, at every horizon.

\subsection{Compression against rollout length}

Skill retained by compressed copies of the 25 five-split parents, on 20 held-out trajectories (descriptive).

\begin{center}\small
\begin{tabularx}{\textwidth}{@{}>{\raggedright\arraybackslash\hsize=1.000\hsize}Xrrrr@{}}
\toprule
\textbf{Compression} & \textbf{1 step} & \textbf{24 steps} & \textbf{100 steps} & \textbf{368 steps} \\
\midrule
Weights rounded to 8 bits & 99.7\% & 99.3\% & 97.0\% & 95\% \\
20\% of weights pruned & 97.4\% & 95\% & 80\% & 57\% \\
30\% of weights pruned & 90\% & 83\% & 39\% & −41\% \\
30\% pruned, then fine-tuned for 500 iterations & 102.3\% & 100.9\% & 102.2\% & 103.9\% \\
\bottomrule
\end{tabularx}
\end{center}

The third row is the clearest case in this paper of a one-step test misleading. A check one step ahead says the pruned model keeps 90\% of its skill. At the 2 s horizon the method uses, it keeps 39\%. By 7.4 s, it is worse than a forecast that predicts no change (Figure 10). Heavier compression fails sooner: with 50\% of weights pruned, or with recurrent units removed, the copy is already below the floor at 100 steps.

The fine-tuned row is measured against the unpruned parent, so values above 100\% mostly reflect the extra training and not the pruning.

\begin{figure}[tbp]
\centering
\includegraphics[width=0.58\linewidth]{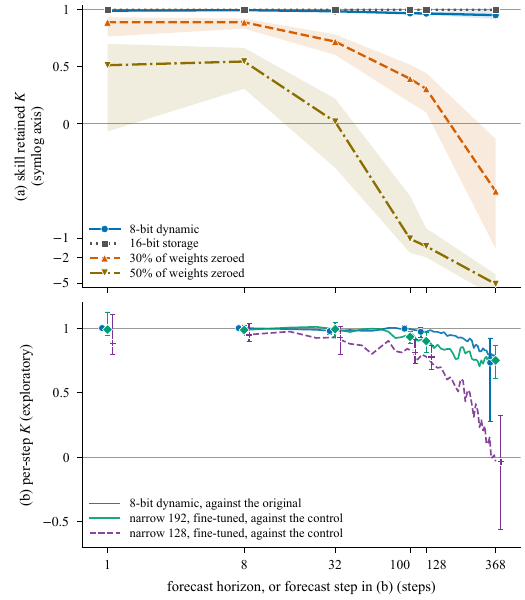}
\caption{(a) Skill retained against the original 32-bit model, by forecast horizon, on 20 held-out trajectories. Bands are 95\% intervals from re-drawing whole trajectories, and the vertical axis is compressed away from zero. (b) Skill retained at every eighth forecast step (exploratory): the 8-bit model against the original, and the fine-tuned narrow models, first seed, against an equally fine-tuned control.}\label{fig:10}
\end{figure}

\subsection{Registered tests}

\begin{itemize}
\tightlist
\item
  \textbf{8-bit quantisation keeps the skill.} PyTorch's dynamic 8-bit quantisation of weights and activations kept 95.2\% {[}91.8\%, 100.9\%{]} of the 368-step skill, above the 90\% margin set in advance. Whether its damage grows with the horizon was not settled. Storing weights in 16 bits cost nothing measurable.
\item
  \textbf{Fine-tuning repairs pruning.} With the pruned weights held at zero, 500 iterations of fine-tuning raised the 100-step skill of the 30\%-pruned copies by 63.9 points {[}54.4, 76.0{]}, from 38\% to 102.2\%. The fine-tuned copies were 99.9\% {[}99.5\%, 100.4\%{]} as skilled as unpruned controls fine-tuned the same way. The same held at 50\% pruning.
\item
  \textbf{Removing recurrent units works at width 192, not clearly at 128.} After fine-tuning, a model cut from 256 to 192 units kept most of the skill of an equally fine-tuned full model at 368 steps: 84\% {[}77\%, 90\%{]} on the first seed, and 90\% {[}84\%, 94\%{]} in a replication on two more. At width 128 the result was not settled.
\item
  \textbf{Pruning is not a reliable route to a better small model.} Against narrow models trained from scratch for the same total number of iterations, pruning was never found better at 368 steps. At width 128, the first seed found scratch training better, and the replication did not settle it. This agrees in part with Liu et al.~\cite{r36}.
\item
  \textbf{Moderate pruning adds error without making the simulator unstable.} At 30\% pruning, the growth rate shifts by +0.03 per second {[}−0.02, +0.07{]}, not separable from zero. Only 7 of 150 pruned or narrowed copies acquired a positive growth rate, all of them at 50\% pruning or reduced width. This is why the growth rate of Section 7 cannot detect pruning damage.
\end{itemize}

\subsection{Cost on a 2017 laptop CPU}

All cost results are exploratory. Times are medians over four rounds, taken after the processor was idle and cooled.

\begin{center}\small
\begin{tabularx}{\textwidth}{@{}>{\raggedright\arraybackslash\hsize=1.300\hsize}X>{\raggedleft\arraybackslash\hsize=0.850\hsize}X>{\raggedleft\arraybackslash\hsize=1.400\hsize}X>{\raggedleft\arraybackslash\hsize=0.700\hsize}X>{\raggedleft\arraybackslash\hsize=0.750\hsize}X@{}}
\toprule
\textbf{Variant} & \textbf{One step, one robot (µs)} & \textbf{One policy update of imagination: 8,192 robots for 24 steps (s)} & \textbf{File size (MiB)} & \textbf{Energy per step (mJ)} \\
\midrule
Model, 32-bit & 953 & 6.11 & 5.02 & 12.7 \\
Model, 8-bit & 1,279 & 5.37 & 1.16 & 15.7 \\
Model, compiled, 32-bit & 672 & 5.70 & 5.02 & 9.61 \\
Model, compiled, 8-bit & 501 & 5.35 & 1.16 & 6.28 \\
Width 192, fine-tuned, compiled, 32-bit & 537 & 3.83 & 2.99 & 7.53 \\
Width 128, fine-tuned, compiled, 32-bit & 439 & 2.36 & 1.49 & 4.99 \\
Five-head ensemble, 32-bit & 1,816 & 9.77 & 7.09 & 24.0 \\
\bottomrule
\end{tabularx}
\end{center}

\begin{figure}[tbp]
\centering
\includegraphics[width=0.62\linewidth]{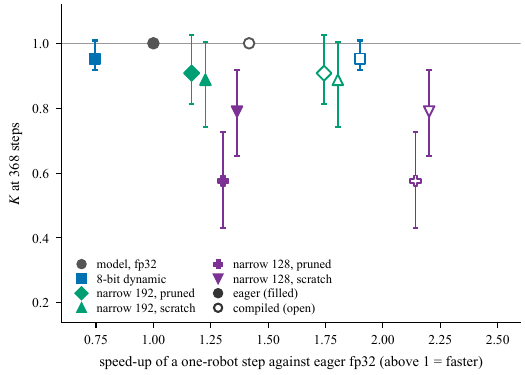}
\caption{Speed against skill on the 2017 laptop CPU (exploratory). Horizontal: speed-up of a one-robot, one-thread step over the ordinary 32-bit model. Vertical: skill retained at 368 steps against the original 32-bit model, first seed, with 95\% intervals. Filled markers are ordinary PyTorch, and open markers are compiled with TorchScript. Models trained from scratch are placed at the speed of the pruned model of the same width. The reference also credits the fine-tuned narrow models with fine-tuning's own gain.}\label{fig:11}
\end{figure}

Three points matter for deployment.

\begin{itemize}
\tightlist
\item
  \textbf{The runtime decides whether 8-bit pays.} In ordinary PyTorch, an 8-bit step takes 1.34 times as long as a 32-bit one. Compiled with TorchScript, it takes 0.75 times as long (Figure 11). At width 128, the compiled 8-bit model is no faster. This CPU has no 8-bit dot-product instruction, and results on such hardware vary~\cite{r52}.
\item
  \textbf{Every form is far faster than real time.} One step of motion lasts 20 ms; the slowest form we timed, the 8-bit ensemble, computes it in under 4 ms on one thread.
\item
  \textbf{The ensemble doubles the cost.} The five-head model takes about twice as long per step as the single network.
\end{itemize}

\subsection{A warning from the released checkpoint}

The same compressions were applied to the released five-head checkpoint for the robot-level study of Section 9. It tolerated them far less well. Rounding its weights to 8 bits raised its 100-step error from 0.40 to 1.21, and pruning 20\% of its weights raised it to 1.32. Both are above the floor of 0.80, so both copies are worse than predicting no change at the horizon of use (descriptive, in-sample). Policies trained in either copy did not walk (Section 9.2).

We did not study why. The released checkpoint was trained on other data, by a run we cannot inspect, and its weights may simply be less redundant. The practical point stands: a compression that is safe for one trained simulator must be re-checked, at the horizon of use, on the simulator that will actually be deployed.

\section{Robot-level evidence: policies trained inside simulators}

Prediction accuracy did not tell us which simulators give walking policies. Policies that walk in our physics model came only from the released model, from its two lightest compressed copies, and from copies of it fine-tuned very briefly on our physics data. Every simulator we trained from scratch gave policies that fall or stand still. The registered tests of whether any check predicts policy quality were not settled, on nine simulators or on nineteen. In the larger test, error one and eight steps ahead ranked the simulators partly in reverse, and running the authors' policy inside each simulator ranked them by walking (secondary).

\subsection{A hand-built physics model of the robot}

The recorded data come from Isaac Lab~\cite{r48}, which needs a GPU we do not have. We therefore built a model of ANYmal D in MuJoCo 3.10~\cite{r44} from the manufacturer's published robot description.

\begin{itemize}
\tightlist
\item
  \textbf{Body.} A free-floating trunk and 12 hinge joints, 57.03 kg in total, with simple collision shapes: a box for the trunk, capsules for the legs and spheres for the feet. Physics runs at 200 Hz and the policy acts at 50 Hz.
\item
  \textbf{Motors.} Isaac Lab drives this robot's joints with a learned actuator network~\cite{r45} that we could not obtain. Following a fallback fixed in advance, we fitted a position controller with a fixed delay to the recorded torques: stiffness 82.6 N m/rad, damping 4.93 N m s/rad, delay 20 ms, R² = 0.93.
\item
  \textbf{Calibration.} Four constants were chosen on a fixed grid of 108 points, using the eight training episodes only, and frozen before any held-out number or policy existed: friction 0.6, contact time constant 0.01 s, joint armature 0.02 kg m² and joint damping 0.
\item
  \textbf{Fidelity.} Replaying the recorded joint commands on the four held-out trajectories, the model beats the floor up to 24 steps (error 0.09 against a floor of 0.16 at one step; 0.23 against 0.64 at 24 steps), and has drifted past it by 100 steps (1.04 against 0.69), as a learned simulator would.
\item
  \textbf{Gate.} A test fixed in advance required that the authors' own released policy walks in the model. It fell in 1 of 80 episodes. Two feet were on the ground 79\% of the time, against 84\% in the recording, and its speed-tracking error was 0.069 m/s, against 0.072.
\end{itemize}

This is a separate physics model in which the authors' policy walks. It is not the simulator that produced the data, and it is not hardware.

\subsection{Nine versions of one simulator (registered)}

Policies were trained with the authors' lightweight code and settings: 4,096 imagined robots, 100 imagined steps per update, 500 updates, and no access to the physics model during training. Each policy was then run in the physics model for 80 episodes that follow the velocity commands of the ten recorded episodes.

A pilot decided the pool. A policy trained in the released model fell in 10\% of episodes. Policies trained in three of our rebuilt single-network simulators fell in every episode, although their imagined returns were higher (5.13, 3.95 and 3.79, against 2.07). Under a rule fixed in advance, the study therefore used the released model and eight compressed copies of it, with two policies per simulator.

The three registered tests were not settled.

\begin{center}\small
\begin{tabularx}{\textwidth}{@{}>{\raggedright\arraybackslash\hsize=1.160\hsize}X>{\raggedright\arraybackslash\hsize=0.840\hsize}Xl@{}}
\toprule
\textbf{Registered question} & \textbf{Estimate} & \textbf{Verdict} \\
\midrule
R0: does the 100-step error rank simulators by physics return? & rank correlation +0.23 {[}−0.87, +0.81{]} & not settled \\\addlinespace[2pt]
R1: growth rate against one-step error & −0.68 {[}−1.50, 0.00{]} & not settled \\\addlinespace[2pt]
R2: kinematic residual against 8-step error & −0.05 {[}−0.47, 0.00{]} & not settled \\
\bottomrule
\end{tabularx}
\end{center}

Per simulator, the outcomes and the checks were as follows. The checks are measured on the recorded data, which the released model may have trained on. The walking scores were measured later, with the same policies.

\begin{center}\footnotesize
\begin{tabularx}{\textwidth}{@{}>{\raggedright\arraybackslash\hsize=1.695\hsize}X>{\raggedleft\arraybackslash\hsize=0.994\hsize}X>{\raggedleft\arraybackslash\hsize=0.643\hsize}X>{\raggedleft\arraybackslash\hsize=1.519\hsize}X>{\raggedleft\arraybackslash\hsize=0.701\hsize}X>{\raggedleft\arraybackslash\hsize=0.701\hsize}X>{\raggedleft\arraybackslash\hsize=0.701\hsize}X>{\raggedleft\arraybackslash\hsize=0.877\hsize}X>{\raggedleft\arraybackslash\hsize=1.169\hsize}X@{}}
\toprule
\textbf{Simulator} & \textbf{Physics return} & \textbf{Falls} & \textbf{Walking score, two policies} & \textbf{$E_{1}$} & \textbf{$E_{8}$} & \textbf{$E_{100}$} & \textbf{Growth rate} & \textbf{Kinematic residual} \\
\midrule
Released model & 16.8 & 37\% & 0.610, 0.573 & 0.083 & 0.093 & 0.40 & 0.70 & 3.3 \\
16-bit weights & 15.2 & 39\% & 0.600, 0.622 & 0.083 & 0.093 & 0.40 & 0.61 & 3.3 \\
10\% pruned & 9.4 & 38\% & 0.249, 0.532 & 0.134 & 0.155 & 0.52 & 0.78 & 3.5 \\
20\% pruned & 3.9 & 19\% & 0.063, 0.061 & 0.219 & 0.298 & 1.32 & 0.80 & 7.0 \\
8-bit weights & 2.7 & 52\% & 0.058, 0.060 & 0.286 & 0.408 & 1.21 & 0.96 & 7.3 \\
30\% pruned & −27.8 & 54\% & 0.004, 0.062 & 0.298 & 0.487 & 1.05 & 0.23 & 7.0 \\
Width 192 & −60.8 & 56\% & 0.051, 0.007 & 0.254 & 0.524 & 1.12 & 0.19 & 7.1 \\
50\% pruned & 9.9 & 0\% & 0.069, 0.069 & 0.370 & 0.958 & 7.56 & 1.80 & 363 \\
Width 128 & 9.7 & 0\% & 0.069, 0.069 & 0.326 & 0.883 & 5.17 & 1.28 & 187 \\
\bottomrule
\end{tabularx}
\end{center}

Four things are worth reading off this table.

\begin{itemize}
\tightlist
\item
  \textbf{The registered outcome rewarded standing still.} Policies trained in the two most damaged simulators never fall, because they stand and follow no command. Their physics return ranks third and fourth of nine. This is why the walking score of Section 3.5 was introduced afterwards: standing still scores 0.069 on it.
\item
  \textbf{By walking score, three simulators give walking policies and six do not}, and the two policies of a simulator agree every time. The three are the released model and its two lightest copies.
\item
  \textbf{Those three are also the three with the lowest prediction errors and the lowest kinematic residual.} The growth rate would have chosen differently: its best three were the width-192 copy, the 30\%-pruned copy and the 16-bit copy. This is a descriptive reading made after the fact.
\item
  \textbf{Even the best policies are immature.} They fell in 31\% to 45\% of episodes, against 1 in 80 for the authors' policy. Five hundred updates is the budget this laptop allows.
\end{itemize}

The pool is nine versions of one simulator that differ only by compression, so it cannot speak about training choices.

\subsection{Why policies from our rebuilt simulators fall}

The pilot's rebuilt simulators are accurate on held-out recordings and useless for policy training. A descriptive diagnosis found why.

\begin{itemize}
\tightlist
\item
  \textbf{The data contain no fall.} Not one of the 10,000 recorded rows is a fall. The network that predicts the end of an episode has nothing to learn from.
\item
  \textbf{The simulators almost never imagine a fall.} In imagination, the single-network simulators ended at most 0.2\% of 512 episodes in a fall. The same policies fell in all 80 episodes in physics. Given the true history, these simulators predicted a real fall within 0.2 s in 0 to 12\% of cases. The released model also missed all 8 such falls of its own policy.
\item
  \textbf{The policies leave the data.} The failed policies' imagined states and commands lie 6.6 to 7.9 times further from the training data than the authors' policy's do. For the policy trained in the released model, the ratio is 1.9.
\item
  \textbf{Few commands are covered.} Only 12.8\% of the velocity commands used in policy training lie near one of the 21 recorded commands.
\item
  \textbf{The uncertainty penalty alone does not fix it.} We rebuilt the model with five heads on the same data. One policy stood still (no falls, speed error 0.80 m/s), and one fell in every episode. In imagination, that simulator predicted a fall in every episode by about step 17.
\end{itemize}

This is model exploitation in a concrete form: reinforcement learning finds the region where the simulator is wrong in the policy's favour, and a recording of flawless walking does not teach the simulator what failure looks like.

\subsection{Simulators trained on 6 million physics transitions}

If the data are the problem, more and broader data should help. We generated 6 million transitions in the MuJoCo model: 55\% from the authors' policy, 25\% from the 22 policies of Section 9.2, and the rest with heavy noise or random motion, with half of the episodes under varied physics. We trained five-head simulators from scratch on all of it and on a 10,000-transition subset, each at two training lengths.

On held-out physics data, these simulators are accurate. At 8 steps, the 6-million simulator's error is 0.097, against 0.39 for the floor, in that simulator's own units. A second gate, fixed in advance, then ran the authors' released policy inside each simulator, and it failed twice. In the 6-million simulator, the policy fell in 15.5\% of imagined episodes and tracked its command 31\% as well as in physics; after training twice as long, 26.6\% and 49\%. The limits were 10\% and 80\%.

The table puts these simulators beside the released model on the same held-out physics data, in common units, so its errors differ from the pair just quoted. The \emph{walk test} runs the authors' policy inside the simulator; its last two columns are the share of imagined episodes that fall, and the tracking reward as a share of the physics model's (descriptive).

\begin{center}\footnotesize
\begin{tabularx}{\textwidth}{@{}>{\raggedright\arraybackslash\hsize=2.046\hsize}X>{\raggedleft\arraybackslash\hsize=0.592\hsize}X>{\raggedleft\arraybackslash\hsize=0.592\hsize}X>{\raggedleft\arraybackslash\hsize=0.592\hsize}X>{\raggedleft\arraybackslash\hsize=0.592\hsize}X>{\raggedleft\arraybackslash\hsize=1.185\hsize}X>{\raggedleft\arraybackslash\hsize=1.400\hsize}X@{}}
\toprule
\textbf{Simulator} & \textbf{$E_{1}$} & \textbf{$E_{8}$} & \textbf{$E_{100}$} & \textbf{$E_{368}$} & \textbf{Walk test: falls} & \textbf{Walk test: tracking share} \\
\midrule
Released model & 0.122 & 0.149 & 0.354 & 0.718 & 0.00 & 0.87 \\
6 million transitions & 0.101 & 0.152 & 0.393 & 5.878 & 0.23 & 0.30 \\
6 million, trained twice as long & 0.087 & 0.124 & 0.280 & 1.616 & 0.39 & 0.47 \\
10,000 transitions & 0.130 & 0.206 & 0.432 & 8.321 & 0.53 & 0.35 \\
10,000, trained twice as long & 0.140 & 0.200 & 0.381 & 8.326 & 0.55 & 0.53 \\
\bottomrule
\end{tabularx}
\end{center}

The third row is the instructive one. It predicts this physics more accurately than the released model at 1, 8 and 100 steps. The released model has never seen this physics, yet the authors' policy walks inside it, and does not walk inside the more accurate simulator. Only the longest horizon and the walk test tell the two apart.

Policies trained in the from-scratch simulators did not walk. In the 6-million simulator, the walking score fell from 0.017 to 0.000 over 2,000 updates, while imagined episodes shrank from 228 steps to 1: the policy learned to end the imagined episode at once, because the penalty for disagreement between the heads outweighed any reward for walking.

Our own review traced part of this to two design mistakes. Our trainer divided each state by its own spread in the data, not by the authors' fixed scales, so its loss weighted states such as body tilt and vertical speed far less than theirs does. And the simulators were trained for 2,572 updates of 256 windows (5,144 when doubled), about 0.1 (0.2) of one pass over the data, while the authors' loop passes over all of its data in every iteration. Rather than repair them, the next study adapted the released model, which has neither problem.

\subsection{The adaptation ladder (registered)}

The working simulator we have is the released model. The final study therefore changes it in small, controlled steps, and asks when its policies stop walking and whether any check sees that coming.

\textbf{Design.} The released model is fine-tuned on MuJoCo data and saved after 5, 20, 40, 80, 160, 320 and 1,280 optimiser steps. Runs differ in the data (walking only, or mixed), the learning rate (the original 1e-4, or 1e-5) and the training windows (rollouts, or one step ahead). Each run tests three saved simulators: first the one saved after 80 steps, then 20 if it failed or 320 if it did not, then one more step count chosen to locate the point where policies stop walking. Only walking scores steer this search; no check does. A simulator is tested with two policies, and a third when they disagree; it fails when most of its policies score below 0.15. The decision rules and the read-only analysis script were committed before any ladder policy was trained.

\textbf{The finished record.} The job ended on 17 November 2026 at 01:51 IST, after 53.0 of at most 56 hours, with 40 policies measured in 18 simulators; 4 of the 18 tests needed a third policy. Figure 12 shows every tested simulator, and Appendix C lists each policy's walking score and fall rate. For comparison, the released model's own two policies score 0.610 and 0.573 (descriptive).

\begin{figure}[tbp]
\centering
\includegraphics[width=1.00\linewidth]{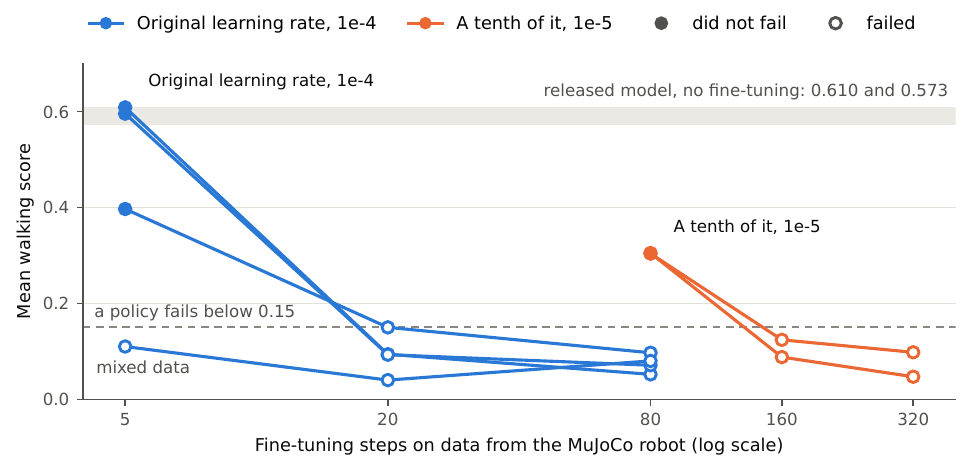}
\caption{Policies stop walking after 20 fine-tuning steps, or 160 at a tenth of the learning rate. The adaptation ladder. Each point is one copy of the released model fine-tuned on data from our MuJoCo robot for the given number of steps, placed at the mean walking score of its two or three policies; lines join copies saved from one run. Filled points: the simulator did not fail. Open points: at least two of its policies failed, so the simulator failed. The open point at 0.15 and 20 steps is the run with one-step windows: its mean rounds to the threshold, but two of its three policies scored below it. The band marks the released model's own two policies. Descriptive.}\label{fig:12}
\end{figure}

A score of 0.000 marks a policy whose training broke down inside its simulator, which the rules count as a failed policy; 3 of the 40 policies broke down this way. A second seed of the mixed-data run was skipped by rule, because the first found no point at which policies still walked.

\textbf{What the ladder shows.} At the original learning rate, all 8 simulators fine-tuned for 20 or 80 steps failed, and 3 of the 4 fine-tuned for 5 steps did not. The exception was the run on mixed data, which failed at all three step counts it tested. With walking data, the point of failure lay between 5 and 20 steps, for both seeds and for the run with one-step windows. At a tenth of the learning rate, both simulators fine-tuned for 80 steps did not fail, and all 4 fine-tuned for 160 or 320 steps failed, with walking or mixed data alike. So 20 fine-tuning steps on data from the target physics at the original learning rate, or 160 at a tenth of it, were enough to stop the released model giving walking policies. Of the 27 failed policies, 21 fell in more than half of their episodes; both policies of the mixed-data simulator fine-tuned for 80 steps barely fell (fall rates 0.06 and 0.00) and still scored low (Appendix C).

\textbf{Short-range error improved while walking got worse.} On held-out episodes of the authors' policy walking in the physics model, all 18 fine-tuned copies had lower 1-step and 8-step error than the released model. Only 2 had lower 100-step error, and only 2 lower 368-step error. Thirteen of the 18 failed (descriptive).

\textbf{The registered analysis.} It ranks 19 simulators, the released model and the 18 tested copies, by each check and by their mean walking score. It compares the two orders with Spearman's rank correlation: +1 when a check orders the simulators exactly as their policies walk, −1 for the reverse, with lower errors and residuals counted as better. Intervals resample the 19 simulators 20,000 times. Neither the premise nor the primary questions were settled.

\begin{center}\small
\begin{tabularx}{\textwidth}{@{}>{\raggedright\arraybackslash\hsize=1.000\hsize}Xrll@{}}
\toprule
\textbf{Question} & \textbf{Estimate} & \textbf{Interval} & \textbf{Verdict} \\
\midrule
Premise: does 100-step error rank the simulators by walking? & +0.16 & 95\% {[}−0.34, +0.65{]} & not settled \\\addlinespace[2pt]
Primary: does the growth rate rank them better than 1-step error? & +0.49 & 97.5\% {[}−0.42, +1.32{]} & not settled \\\addlinespace[2pt]
Primary: does the kinematic residual rank them better than 8-step error? & +0.75 & 97.5\% {[}−0.17, +1.42{]} & not settled \\\addlinespace[2pt]
Secondary: does 368-step error rank them better than 8-step error? & +0.75 & 95\% {[}−0.01, +1.35{]} & not settled \\\addlinespace[2pt]
Secondary: walk test, tracking reward, against 8-step error & +1.22 & 95\% {[}+0.71, +1.58{]} & walk test better \\\addlinespace[2pt]
Secondary: walk test, disagreement between the heads, against 8-step error & +1.26 & 95\% {[}+0.74, +1.64{]} & walk test better \\\addlinespace[2pt]
Secondary: walk test, share of imagined falls, against 8-step error & +0.72 & 95\% {[}−0.30, +1.60{]} & not settled \\
\bottomrule
\end{tabularx}
\end{center}

The first row is a rank correlation. The others are differences between two rank correlations, the first check's minus the second's, so they can run from −2 to +2. The output was written on 17 November; a rerun of the unedited, registered script reproduces it byte for byte.

\textbf{What the analysis shows.} Figure 13 shows how each check ranked the simulators. Error one and eight steps ahead ranked them partly in reverse, because fine-tuning lowered that error while it stopped the policies walking. The large differences in favour of the label-free checks owe much to that reversal, and with 19 related simulators the intervals stayed too wide to settle either primary question.

\begin{figure}[tbp]
\centering
\includegraphics[width=1.00\linewidth]{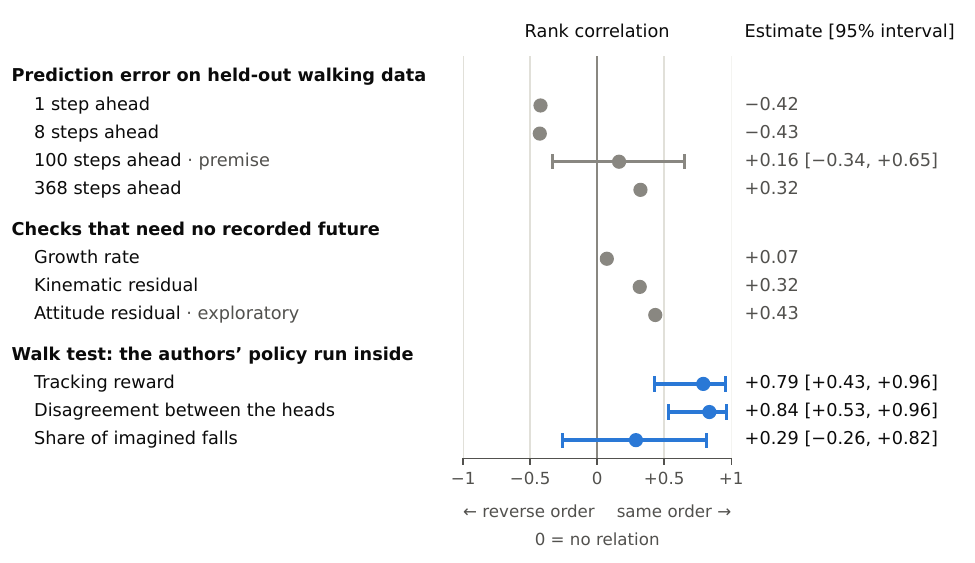}
\caption{The walk test ranked the 19 simulators by walking best, at +0.84 and +0.79. Error one and eight steps ahead ranked them partly in reverse. How each check ranked the ladder's 19 simulators, the released model and 18 fine-tuned copies of it, against the mean walking score of their policies, as Spearman's rank correlation. Lower errors, residuals, growth rate, imagined falls and disagreement count as better, so a point right of zero means the check orders the simulators as their policies walk. Whiskers are 95\% intervals; the registered analysis computed them only for these four rankings, and for the others only for the differences in the table above. The walk-test rows are secondary results; the attitude residual was registered as exploratory, without a decision rule.}\label{fig:13}
\end{figure}

\textbf{The walk test ranked them well.} Running the authors' policy inside each simulator ordered the 19 by walking: both its tracking reward and the disagreement between the five heads ranked them better than 8-step error did (secondary). The tracking shares differed by only a few percentage points between simulators. Imagined falls were too rare in every simulator to rank them.

\textbf{Choosing by error would have chosen failures.} Asked for the three best simulators, 1-step and 8-step error each picked three copies fine-tuned on walking data, and all of them failed. The best three by walking score average 0.60; the picks of 1-step and 8-step error fell short of that by 0.53 and 0.54 (95\% intervals {[}0.27, 0.55{]} for both). The 368-step error and the kinematic residual both picked two copies fine-tuned for 5 steps and the released model, which fell short by only 0.07 ({[}0.00, 0.33{]} and {[}0.00, 0.34{]}; secondary).

\textbf{Other secondary results.} No check separated the failed simulators from the others with an interval clear of chance. The kinematic residual ranked the simulators better than 8-step error in four variants: with the eight compressed copies of Section 9.2 added (27 simulators), on a mixed test set, and with the fall rate or the physics return as the outcome. In the 27-simulator pool, the premise held, and 1-step and 8-step error no longer ranked in reverse, probably because most compressed copies have both large errors and failing policies. Appendix A lists these results.

\textbf{How to read these results.} Four cautions apply; the registration names three of them.

\begin{itemize}
\tightlist
\item
  \textbf{The intervals are optimistic.} The search put its tests near the point of failure on purpose, and the bootstrap treats the tested simulators as a random sample.
\item
  \textbf{A reversed baseline flatters every comparison with it.} The label-free checks and the walk test were each compared with an error check that ranked the simulators partly backwards.
\item
  \textbf{One cause may drive everything.} How far a simulator was moved from the released model may move every check and the walking score together. The walk test, and the heads' disagreement above all, may measure little more than that distance.
\item
  \textbf{One lineage.} All 19 simulators descend from the released model, so none of this shows how the checks rank simulators built another way.
\end{itemize}

\subsection{What the robot-level evidence shows}

\begin{itemize}
\tightlist
\item
  \textbf{Walking policies have come from one lineage only.} Of the 33 simulators in which a policy has been trained (nine in Section 9.2, four rebuilt ones in Sections 9.2 and 9.3, two trained from scratch in Section 9.4 and 18 in the ladder), the ones whose policies walk are the released model, two lightly compressed copies of it, and the five ladder simulators that did not fail, fine-tuned for 5 steps at the original learning rate or 80 at a tenth of it. Nothing we trained from scratch, on the recording or on 6 million physics transitions, is among them.
\item
  \textbf{Open-loop accuracy did not separate them.} A from-scratch simulator beat the released model on held-out physics data at 1, 8 and 100 steps, and still failed the walk test: the authors' policy fell in 39\% of imagined episodes inside it, and in none inside the released model. Of the two from-scratch simulators in which we trained policies, the one trained on 6 million transitions beat the released model one step ahead, and neither gave a policy that walks. What a policy needs is a simulator that behaves sensibly where the policy wanders during training, and an error measured along recorded commands does not test that. On the ladder, every fine-tuned copy beat the released model one and eight steps ahead, and 13 of the 18 failed.
\item
  \textbf{A closed-loop test ranked them best.} In the record that preceded the ladder, the walk test's tracking share was 0.81 or more for the three simulators of Section 9.2 whose policies walk, and 0.59 or less for the six whose policies do not. It was also low for every simulator we trained from scratch. That record was descriptive, and known before the ladder was registered. On the ladder's 19 simulators, the walk test's tracking reward and the heads' disagreement ranked them by walking better than 8-step error did (registered secondary tests, Section 9.5). The walk test may measure only how far a simulator has moved from the one the trusted policy was trained in, and it needs a trusted policy, which a team building its first simulator may not have.
\item
  \textbf{Limits.} One robot and one gait, a physics model we built, policies trained for 500 updates, a walking score with a threshold we chose, and simulators that all descend from one released checkpoint.
\end{itemize}

\section{What we got wrong, and what caught it}

Nine numbered claims of this project were withdrawn on evidence, each by a check the project ran on itself. We report them because most are mistakes that an evaluation of a robot model can make without anyone noticing, and because several of the results above exist only because an earlier version was wrong.

\subsection{The mistakes}

\begingroup\small
\setlength{\LTpre}{6pt}\setlength{\LTpost}{6pt}
\begin{xltabular}{\textwidth}{@{}>{\raggedright\arraybackslash\hsize=0.800\hsize}X>{\raggedright\arraybackslash\hsize=1.250\hsize}X>{\raggedright\arraybackslash\hsize=0.950\hsize}X@{}}
\toprule
\textbf{What went wrong} & \textbf{How much it mattered} & \textbf{What we do now} \\
\midrule
\endfirsthead
\toprule
\textbf{What went wrong} & \textbf{How much it mattered} & \textbf{What we do now} \\
\midrule
\endhead
\bottomrule
\endlastfoot
\textbf{Overlapping trajectories were counted as independent.} & 3,200 overlapping trajectories came from 4 independent ones. The worst 5\% carried 92.6\% of the total error, and all of it came from two short stretches of data sampled again and again. & Report the number of independent units beside every interval. \\\addlinespace[2pt]
\textbf{A headline depended on how a metric was averaged.} & `The released checkpoint loses to predicting no change' held under one of four ways of aggregating the error, and failed on non-overlapping trajectories. Its own gating checks withdrew it. & Fix the aggregation in advance, and look at each state dimension. \\\addlinespace[2pt]
\textbf{A decision rule was tied to the wrong horizon.} & The central claim went from `settled' to `cannot be settled' in all eight settings when re-measured, and was confirmed only at the horizon the claim is about (Section 5.2). & Tie each rule to the horizon of use, before the runs. \\\addlinespace[2pt]
\textbf{A rule was called pre-registered when it was not.} & It reached version control 2.9 hours after the runs it governed had finished. Its claim to registration was withdrawn. & A rule counts only if it is committed before the data exist. Saying it in conversation does not count. \\\addlinespace[2pt]
\textbf{Intervals resampled the wrong unit.} & Pooling seeds with trajectories and resampling the pool narrowed intervals by a mean factor of 1.42, and changed 1 of 16 verdicts. & Resample whole trajectories, with seeds pooled inside each draw. \\\addlinespace[2pt]
\textbf{A test treated the 45 state numbers as independent.} & A binomial test overstated one result's significance by a factor of about $10^{13}$. & Permute whole trajectories, which keeps the state dimensions together. \\\addlinespace[2pt]
\textbf{A reproducibility figure was inflated.} & A checker counted files that a fresh copy already contained as regenerated, a 50-fold overstatement. & Count only files on an explicit list of regenerated outputs. \\\addlinespace[2pt]
\textbf{Numbers typed by hand went stale.} & Four times, including a percentage that disagreed with itself across two files. & Papers are generated; a build stops on any number not read from a results file. \\\addlinespace[2pt]
\textbf{A robot-level outcome rewarded standing still.} & Two simulators whose policies never move ranked third and fourth of nine (Section 9.2). & A walking score that gives standing still almost nothing. \\\addlinespace[2pt]
\textbf{Our own simulators were built with avoidable flaws.} & Rebuilt single-network simulators had no uncertainty penalty; the from-scratch simulators used the wrong state scaling and too little training (Sections 9.3 and 9.4). & Fix one design difference at a time against the working release. \\
\end{xltabular}
\endgroup

\subsection{The tooling that caught them}

\begin{itemize}
\tightlist
\item
  \textbf{A claims ledger with permanent numbers.} Every claim has an identifier that is never edited in place. A claim that turns out wrong is marked superseded, points to what replaced it, and stays. The rebuild's ledger holds 310 entries, 24 of them superseded.
\item
  \textbf{Decision rules committed before the data.} A rule states its statistic, its threshold and its possible verdicts, including `not settled'. Many of the verdicts in this paper are `not settled', and we report those as plainly as the others.
\item
  \textbf{Differential testing against the release.} Running the authors' released checkpoint through our own code is what exposed the one-step-late command (Section 4.3). Internal consistency checks would never have found it.
\item
  \textbf{Generated papers.} Every earlier paper of the project is built from a template whose numbers are read from result files. This paper is the exception: its numbers were copied from those files and checked against them by hand, and are not yet bound to them by a build script.
\item
  \textbf{Independent re-computation.} Statistics of the later studies were recomputed with separately written code, and each draft was read by an AI reviewer that had not seen how it was produced.
\end{itemize}

\subsection{What this cost and what it bought}

The retractions cost time, and they shrank several claims. They also produced the paper's most transferable findings. The count of independent trajectories (four, for the main split) bounds every long-horizon claim made about this data set, including the ten sampled trajectories of the released evaluation script. The off-by-one command, the collapsed variance and the reversal at one step were all found while trying to explain a number that did not fit.

\section{Discussion: working rules for using a learned simulator}

For a team that wants to train a legged-robot policy inside a learned simulator, our results reduce to ten working rules. Each is supported for one robot and one gait. The first eight rest on settled results; the last two rest on secondary or descriptive evidence.

\begingroup\small
\setlength{\LTpre}{6pt}\setlength{\LTpost}{6pt}
\begin{xltabular}{\textwidth}{@{}>{\raggedright\arraybackslash\hsize=1.000\hsize}X>{\raggedright\arraybackslash\hsize=1.000\hsize}Xl@{}}
\toprule
\textbf{Rule} & \textbf{Why} & \textbf{Section} \\
\midrule
\endfirsthead
\toprule
\textbf{Rule} & \textbf{Why} & \textbf{Section} \\
\midrule
\endhead
\bottomrule
\endlastfoot
1. Read the recording before the paper. Find episode boundaries, resets and the timing of commands from the numbers themselves. & Both defects of the released pipeline were visible in the data and in no document. & 3.2, 4.3 \\\addlinespace[2pt]
2. Verify a reimplementation against the release down to losses and gradients, and run the released checkpoint through your own code. & Three loss terms were inactive and one command arrived a step late; neither shows in the configuration. & 4.1, 4.2 \\\addlinespace[2pt]
3. Train the dynamics model on its own rollouts of eight steps or more. & It beat one-step training at long horizons on every held-out episode, and it is what makes the simulator damp its own errors. & 5 \\\addlinespace[2pt]
4. Measure a simulator's prediction error, and that of every change to it, eight or more steps ahead on held-out episodes. Never one step ahead, and never as a stand-in for how its policies walk. & One-step error ranked trained simulators backwards and hid the damage of pruning. Neither 1-step nor 8-step error ranked fine-tuned simulators by their policies. & 5.3, 7, 8, 9.5 \\\addlinespace[2pt]
5. Use the error growth rate as a screen, never as a certificate. & A positive rate flagged one-step-trained simulators. A negative rate certified nothing. & 7.4 \\\addlinespace[2pt]
6. Treat ensemble disagreement as an ordering. If you need a size, rescale it per horizon on held-out episodes, and do not trust its tails. & It ranked error well, and was too small by a factor of 6.6 at one step and 42 at 368 steps. & 6 \\\addlinespace[2pt]
7. Do not build run-time alarms from consistency checks. Re-synchronise on a schedule, and enforce only relations that are exact. & Alarms scored at chance, triggered re-synchronisation lost to a schedule, and an approximate constraint multiplied the error by 2.3. & 7.5 \\\addlinespace[2pt]
8. Compress, then re-check at the horizon of use on the model you will deploy. Fine-tune after pruning. & 8-bit weights were safe on our models and not on the released one. Fine-tuning restored pruned models fully. & 8 \\\addlinespace[2pt]
9. Before training a policy, run a trusted policy inside the simulator and compare with physics. & It ranked 19 fine-tuned simulators by their policies better than 8-step error did (registered secondary tests). It may measure only how far a simulator has moved from the one the trusted policy was trained in. & 9.4 to 9.6 \\\addlinespace[2pt]
10. Treat a fine-tuned simulator as a new simulator, and test its policies again. & Twenty optimiser steps on data from the target physics at the original learning rate, or 160 at a tenth of it, stopped the released model from giving walking policies, while lowering its error one and eight steps ahead. Descriptive. & 9.5 \\
\end{xltabular}
\endgroup

\subsection{Why would a more accurate simulator give worse policies?}

We do not know. The registered analysis of Section 9.5 did not settle it, but it sharpened the question: every fine-tuned copy predicted better than the released model one and eight steps ahead, and most gave worse policies. Three observations constrain the answer.

\begin{itemize}
\tightlist
\item
  \textbf{Policy training is adversarial to the simulator.} The optimiser searches for behaviour that the simulator rewards. The policies that failed had wandered 6.6 to 7.9 times further from the training data than the authors' policy does (Section 9.3). Error measured along recorded commands says nothing about that region.
\item
  \textbf{The simulator must know what failure looks like.} A recording without a single fall produced simulators that almost never predict one. The released model was trained on far more data than was released, by a process we cannot see.
\item
  \textbf{The uncertainty penalty is a scale, and scales do not transfer.} While the authors' policy walks inside it, the released model's heads disagree by 0.46 on average. Our from-scratch simulators disagree more, by 1.08 to 2.04, and with the same penalty weight their policies learned to end the episode instead of walking. Section 6 showed that this scale is not calibrated to begin with. On the ladder, the heads' disagreement while the authors' policy walks stayed at the released model's level in the copies fine-tuned for 5 steps on walking data, rose in all the others, and ranked the simulators by walking better than any error did (Section 9.5). One reading is mechanical: policy training penalises this kind of disagreement, so a simulator whose heads disagree more makes walking costlier to learn. We have not tested this reading.
\end{itemize}

The common thread is the mismatch between how a simulator is scored and how it is used~\cite{r11}. Open-loop error on recorded commands is a necessary check and, on this evidence, not a sufficient one.

\subsection{What transfers beyond this robot}

The numbers are specific to ANYmal D, one gait and this data set. Three things should carry over, because they follow from how such models are built and not from the robot. First, the reversal between one-step and long-horizon ranking follows from feeding predictions back, and we saw it in three of four architectures. Second, the variance collapse follows from the loss alone. Third, the count of independent trajectories is a property of any short recording, and it bounds what can be claimed from it.

\section{Limitations}

\textbf{One robot, one gait, simulated data.} Everything concerns ANYmal D trotting on flat ground, with body measurements only. The recording is 200 s of simulated walking with about twenty velocity commands and no fall. Nothing was run on hardware.

\textbf{Scale.} Our models train on 7,991 transitions, 0.13\% of the 6 million that RWM reports. A training recipe can behave differently at full scale. The released checkpoint, trained at that scale, does not fit our stability account: it amplifies small errors and is still accurate.

\textbf{The released checkpoint is never held out.} Its training rows are not recorded, so every number about it may be in-sample. This includes the uncertainty table of Section 6.3 and all checks of Section 9.2.

\textbf{Few independent trajectories.} The main pool rests on four independent 400-step trajectories, and the five-split pool on twenty. Many registered verdicts are `not settled' for that reason alone. Simulators in a pool share parents and data, so intervals over models are optimistic (Section 7.4).

\textbf{The two label-free checks are narrow.} The growth rate is one leading exponent over a finite window. The kinematic residual tests one relation, between joint angles and joint speeds.

\textbf{The physics model is ours.} It is a hand-built MuJoCo model with a fitted position controller in place of the actuator network, calibrated on the same recording. It is not Isaac Lab and not the robot. Its only external anchor is that the authors' policy walks in it.

\textbf{The policies are immature.} Five hundred updates give policies that still fall often: the best policies of Section 9.2 fell in 31\% to 45\% of episodes. Two or three policy seeds per simulator are few. The walking score and its threshold of 0.15 were chosen by us after the first robot-level study.

\textbf{The robot-level sample is small and of one lineage.} Section 9.2 has nine versions of one simulator. The ladder of Section 9.5 has 18 simulators (its registration allowed up to 21), all descended from the released checkpoint, and its registered premise and primary tests were not settled; the walk test's good ranking there is a secondary result. That registration names its own weaknesses: the search placed its tests near the point of failure on purpose, so bootstrap intervals over its simulators will be optimistic; each simulator's score rests on two or three policies; and how far a simulator was moved from the released model may move every check, the walk test included, and the walking score together.

\textbf{Registration is self-timestamped.} The rules are commits in our own repositories. A third-party archive holds the rebuild study's history; the later studies have none yet. The ladder's analysis output was found on disk with no recorded writer, although a rerun of the unedited registered script reproduces it byte for byte.

\textbf{Cost numbers come from one old laptop.} Its processor lacks the 8-bit instructions of newer hardware, so the speed of 8-bit models will differ elsewhere.

\textbf{One author, with AI assistance.} No independent group has reproduced these results. The code, the model weights and the result files are public so that one can.

\section{Conclusion and future work}

We audited a published learned simulator for a legged robot, from its data file to policies walking in physics, without a GPU. The published training recipe holds up: training on the model's own rollouts is what makes it accurate over the horizons a policy uses, and what makes it damp its own errors. The published uncertainty orders errors correctly and sizes them wrongly. A one-step check, the most common test of such models, misjudged the training recipe, the compression and the uncertainty alike.

The robot-level results are the most sobering. We could verify, measure, compress and diagnose the released model in detail, and we could build a simulator that predicts held-out physics data better than it does up to 100 steps. We could not build one that gives a walking policy. Twenty fine-tuning steps on data from our physics model were enough to stop the released model giving one, while making it more accurate one and eight steps ahead. On this evidence, a learned simulator should be accepted by what policies trained in it do, with open-loop error as a first filter only. Running a trusted policy inside the simulator was the best cheap check we found, and it has yet to be tested on simulators of other origins.

\textbf{Future work, in order of what it would settle.}

\begin{enumerate}
\def\labelenumi{\arabic{enumi}.}
\tightlist
\item
  \textbf{Test the walk test on simulators of other origins.} Register it as a primary test on simulators that do not descend from the trusted policy's own, to learn whether it predicts walking or only measures distance from the released model.
\item
  \textbf{Repeat the robot-level tests in the simulator that produced the data}, with the actuator network and a GPU, to remove our physics model as an explanation.
\item
  \textbf{Train the rebuild at full scale}, on millions of transitions that include falls and recoveries, to test whether the stability account and the policy failures survive.
\item
  \textbf{Turn the closed-loop test into a pre-deployment check} that does not need a trusted policy, starting from the link between imagined return and error (Section 7.4).
\item
  \textbf{Fix the uncertainty at its three sources} (independent members, per-member rollouts and a per-horizon scale), then test whether a calibrated penalty changes which policies walk, starting with the ladder copies that failed.
\item
  \textbf{Train mature policies.} Five hundred updates was the limit of the hardware, not of the method.
\item
  \textbf{Other robots, gaits and terrain, and finally hardware.} Nothing here should be assumed to transfer until it is measured.
\end{enumerate}

\textbf{Two things only the original authors hold} would settle several of these points at once. Knowing which data trained the released checkpoint would tell us which of its measurements are held out (Section 4.4). The 6-million-transition dataset used for pretraining would let us train the rebuild at the published scale (item 3), and repeat the checks of Sections 5 to 7 on the data the method was designed for.

\section*{Code and data availability}
\addcontentsline{toc}{section}{Code and data availability}

Everything needed to check the results is public, except where stated.

\begin{center}\small
\begin{tabularx}{\textwidth}{@{}>{\raggedright\arraybackslash\hsize=1.050\hsize}X>{\raggedright\arraybackslash\hsize=0.950\hsize}X@{}}
\toprule
\textbf{What} & \textbf{Where} \\
\midrule
Rebuild of the model, claims ledger and result files (Sections 3 to 6) & \href{https://github.com/joyjeet-singh/rwm}{\nolinkurl{github.com/joyjeet-singh/rwm}}, Apache 2.0; its history is archived at Software Heritage \\\addlinespace[2pt]
Rebuild checkpoints, with a model card & \href{https://huggingface.co/Joyjeetsingh/rwm-reproduction}{\nolinkurl{huggingface.co/Joyjeetsingh/rwm-reproduction}} \\\addlinespace[2pt]
Comparison of checks, fine-tuning study, other architectures, physics model and robot-level studies, with their registrations (Sections 7 to 9) & \href{https://github.com/joyjeet-singh/rwm-robot-level}{\nolinkurl{github.com/joyjeet-singh/rwm-robot-level}}, Apache 2.0 \\\addlinespace[2pt]
Simulators, policies and training records of Sections 7 to 9, with a SHA-256 manifest & \href{https://huggingface.co/Joyjeetsingh/rwm-robot-level}{\nolinkurl{huggingface.co/Joyjeetsingh/rwm-robot-level}} \\\addlinespace[2pt]
Preprint of the rebuild and uncertainty study & \href{https://arxiv.org/abs/2611.02699}{\nolinkurl{arXiv:2611.02699}}~\cite{r3} \\\addlinespace[2pt]
Error-growth, kinematic-check and compression studies (Sections 5.4, 7.5 and 8) & available from the author on request \\
\bottomrule
\end{tabularx}
\end{center}

\textbf{Inputs we did not create.} The authors' lightweight code, data file and checkpoint (robotic\_world\_model\_lite, Apache 2.0) and their training library (rsl\_rl\_rwm, BSD 3-Clause) are read at pinned commits, and are never modified or redistributed. The robot description is ANYbotics' anymal\_d\_simple\_description (BSD 3-Clause). The physics engine is MuJoCo 3.10.0.

\textbf{Rerunning.} For the rebuild, \texttt{./setup.sh} fetches the pinned upstream code, and \texttt{./reproduce.sh\ -\/-quick\ -\/-force} reruns everything except training. Training stages need the weights, which are published separately. Each later study lists its commands in its own README.

\textbf{Software and hardware.} Training and policy work: Python 3.11 and PyTorch 2.2.2 on a 2017 Intel Core i5-7360U laptop with two cores, 8 GiB of memory and no GPU. Analysis-only studies: a two-core Linux workspace with a newer PyTorch, re-checked on the laptop. The longest single job was the adaptation ladder: 53.0 hours of a 56-hour limit.

\section*{Acknowledgement}
\addcontentsline{toc}{section}{Acknowledgement}

We thank the authors of RWM and RWM-U for releasing their code, data and checkpoint, and the first author for answering our questions. Without the release, none of this could have been checked.

\appendix

\section{Registered verdicts not tabulated in the text}\label{app:A}
Sections 7.3, 9.2 and 9.5 list the registered verdicts of the comparison of checks, the physics test and the ladder. The table below collects the other registered results this paper relies on. Verdict words are the ones each rule fixed in advance. The ladder rows are secondary analyses; unless a row says otherwise, each gives, in order, the premise and the two primary comparisons of Section 9.5 on another pool, test set or outcome.

\begingroup\small
\setlength{\LTpre}{6pt}\setlength{\LTpost}{6pt}
\begin{xltabular}{\textwidth}{@{}>{\raggedright\arraybackslash\hsize=0.600\hsize}X>{\raggedright\arraybackslash\hsize=1.300\hsize}X>{\raggedright\arraybackslash\hsize=1.300\hsize}X>{\raggedright\arraybackslash\hsize=0.800\hsize}X@{}}
\toprule
\textbf{Study} & \textbf{Question} & \textbf{Estimate} & \textbf{Verdict} \\
\midrule
\endfirsthead
\toprule
\textbf{Study} & \textbf{Question} & \textbf{Estimate} & \textbf{Verdict} \\
\midrule
\endhead
\bottomrule
\endlastfoot
Rebuild & Rollout training against teacher forcing at 8 steps & not resolved in any of the eight combinations examined & cannot be settled at this budget \\\addlinespace[2pt]
Rebuild & The same at 368 steps, held-out pair & +1.20 {[}+0.56, +2.05{]}, 10 of 10 episodes positive & reproduces at long horizon \\\addlinespace[2pt]
Rebuild & The same across five splits & +2.2 {[}+1.4, +3.5{]}; $p = 0.002$ over episodes & replicates across folds \\\addlinespace[2pt]
Rebuild & Does the variance collapse on data with known noise? & head 22 times below the true noise & objective-driven \\\addlinespace[2pt]
Rebuild & Disagreement against predicted step size, 100 steps & +0.28 {[}+0.20, +0.39{]} & disagreement beats step size \\\addlinespace[2pt]
Rebuild & One multiplier per horizon, scored on an unseen episode & one-sigma coverage in band at 6 of 6 horizons & within band \\\addlinespace[2pt]
Rebuild & Is the one-step shortfall a few bad channels? & spread across dimensions 10.0 points, against a limit of 15 & uniform \\\addlinespace[2pt]
Error growth, first study & Teacher forcing minus rollout training, growth rate & +1.18 per second {[}1.07, 1.30{]} & teacher forcing more expansive \\\addlinespace[2pt]
Error growth, first study & Growth rate against 368-step error, 55 models & 0.66 {[}0.46, 0.79{]} & predicts \\\addlinespace[2pt]
Error growth, first study & The same within 40 rollout-trained models & 0.21 {[}−0.13, 0.50{]} & not settled \\\addlinespace[2pt]
Error growth, full study & Growth rate against 368-step error, 48 rollout-trained models & 0.85 {[}0.70, 0.93{]} & predicts \\\addlinespace[2pt]
Error growth, full study & Growth signal as a run-time alarm & AUROC 0.50, against 0.64 for a step counter & worse than counter \\\addlinespace[2pt]
Error growth, full study & Training horizon against growth rate & −0.97 {[}−0.99, −0.89{]} & longer is more stable \\\addlinespace[2pt]
Kinematic checks & Re-synchronise when the score fires, or on a schedule? & +10.1\% {[}5.6\%, 15.7\%{]} error for the triggered version & schedule better \\\addlinespace[2pt]
Compression & 8-bit quantisation, skill kept at 368 steps & 95.2\% {[}91.8\%, 100.9\%{]} & keeps \\\addlinespace[2pt]
Compression & Does 8-bit damage grow with the horizon? & interval spans zero & not settled \\\addlinespace[2pt]
Compression & Width 192 after fine-tuning, skill kept at 368 steps & 84\% {[}77\%, 90\%{]}; replication 90\% {[}84\%, 94\%{]} & keeps most \\\addlinespace[2pt]
Compression & Width 128 after fine-tuning & 53\% {[}40\%, 65\%{]}; replication 44\% {[}4\%, 65\%{]} & not settled \\\addlinespace[2pt]
Compression & Pruned against trained from scratch, width 128 & −13.2\% {[}−21.2\%, −5.4\%{]} on the first seed; replication unsettled & scratch better, then not settled \\\addlinespace[2pt]
Fine-tuning & Does fine-tuning restore a 30\%-pruned model at 100 steps? & +63.9 points {[}54.4, 76.0{]} & restores \\\addlinespace[2pt]
Fine-tuning & Fine-tuned pruned copy against fine-tuned control & 99.9\% {[}99.5\%, 100.4\%{]} & as skilled as the control \\\addlinespace[2pt]
Adaptation ladder & 27 simulators, with the compressed copies of Section 9.2: premise; growth rate against 1-step error; residual against 8-step error & +0.47 {[}+0.08, +0.78{]}; +0.11 {[}−0.52, +0.76{]}; +0.44 {[}+0.04, +0.88{]} & supported; not settled; label-free better \\\addlinespace[2pt]
Adaptation ladder & Mixed test set: the same three & 0.00 {[}−0.42, +0.51{]}; +0.65 {[}−0.04, +1.26{]}; +0.76 {[}+0.05, +1.32{]} & not settled; not settled; label-free better \\\addlinespace[2pt]
Adaptation ladder & Fall rate as the outcome: the same three & +0.07 {[}−0.45, +0.53{]}; +0.41 {[}−0.42, +1.16{]}; +0.99 {[}+0.24, +1.45{]} & not settled; not settled; label-free better \\\addlinespace[2pt]
Adaptation ladder & Physics return as the outcome: the same three & +0.25 {[}−0.31, +0.75{]}; +0.84 {[}+0.08, +1.45{]}; +1.16 {[}+0.50, +1.57{]} & not settled; label-free better; label-free better \\\addlinespace[2pt]
Adaptation ladder & Seed-0 runs and the released model (16 simulators): the two primary comparisons & +0.60 {[}−0.32, +1.38{]}; +0.61 {[}−0.32, +1.37{]} & not settled; not settled \\\addlinespace[2pt]
Adaptation ladder & Each simulator's seed-1 policy alone as the outcome: the same three & +0.14 {[}−0.35, +0.64{]}; +0.28 {[}−0.65, +1.10{]}; +0.61 {[}−0.29, +1.33{]} & not settled (all three) \\\addlinespace[2pt]
Adaptation ladder & Does a check separate failed simulators from the others? Chance that a simulator that did not fail has the better value & 1-step error 0.27, 8-step 0.31, 24-step 0.40, 100-step 0.67, 368-step 0.67, growth rate 0.46, residual 0.67; every interval contains 0.5 & not settled (all seven) \\
\end{xltabular}
\endgroup

\section{Dated milestones of registration}\label{app:B}
The dates on which registrations were committed, data were frozen and the main jobs finished. The commits that record them are in the repositories listed under Code and data availability, or available on request.

\begin{center}\small
\begin{tabularx}{\textwidth}{@{}l>{\raggedright\arraybackslash\hsize=1.000\hsize}X@{}}
\toprule
\textbf{Date (2026)} & \textbf{Milestone} \\
\midrule
21 August & History of the rebuild archived by a third party \\\addlinespace[2pt]
28 October & Kinematic-check study registered \\\addlinespace[2pt]
1 November & Comparison of checks registered, then measured \\\addlinespace[2pt]
7 November & Physics model calibrated and frozen \\\addlinespace[2pt]
14 November & Adaptation ladder registered; first ladder policy measured \\\addlinespace[2pt]
17 November & Ladder finished after 53.0 hours: 40 policies in 18 simulators. Its registered analysis was run the same day. \\
\bottomrule
\end{tabularx}
\end{center}

\section{The adaptation ladder, policy by policy}\label{app:C}
Every simulator the ladder tested, with its verdict, its mean walking score (the outcome of the registered analysis), and each policy's walking score with its fall rate in brackets. A policy fails when its walking score is below 0.15, or when its training broke down inside its simulator (score 0.000); a simulator fails when at least two of its policies fail. Descriptive.

\begingroup\small
\setlength{\LTpre}{6pt}\setlength{\LTpost}{6pt}
\begin{xltabular}{\textwidth}{@{}>{\raggedright\arraybackslash\hsize=1.500\hsize}X>{\raggedleft\arraybackslash\hsize=0.750\hsize}X>{\raggedright\arraybackslash\hsize=0.700\hsize}X>{\raggedleft\arraybackslash\hsize=0.750\hsize}X>{\raggedright\arraybackslash\hsize=1.300\hsize}X@{}}
\toprule
\textbf{Run} & \textbf{Fine-tuning steps} & \textbf{Verdict} & \textbf{Mean walking score} & \textbf{Each policy: walking score (fall rate)} \\
\midrule
\endfirsthead
\toprule
\textbf{Run} & \textbf{Fine-tuning steps} & \textbf{Verdict} & \textbf{Mean walking score} & \textbf{Each policy: walking score (fall rate)} \\
\midrule
\endhead
\bottomrule
\endlastfoot
Walking data, 1e-4, rollout windows, seed 0 & 5 & did not fail & 0.609 & 0.509 (0.45), 0.710 (0.18) \\\addlinespace[2pt]
Walking data, 1e-4, rollout windows, seed 0 & 20 & failed & 0.094 & 0.086 (0.73), 0.101 (0.59) \\\addlinespace[2pt]
Walking data, 1e-4, rollout windows, seed 0 & 80 & failed & 0.052 & 0.060 (0.82), 0.045 (0.79) \\\addlinespace[2pt]
Walking data, 1e-4, rollout windows, seed 1 & 5 & did not fail & 0.596 & 0.751 (0.12), 0.441 (0.54) \\\addlinespace[2pt]
Walking data, 1e-4, rollout windows, seed 1 & 20 & failed & 0.093 & 0.084 (0.90), 0.102 (0.59) \\\addlinespace[2pt]
Walking data, 1e-4, rollout windows, seed 1 & 80 & failed & 0.071 & 0.065 (0.72), 0.077 (0.65) \\\addlinespace[2pt]
Walking data, 1e-4, one-step windows, seed 0 & 5 & did not fail & 0.397 & 0.600 (0.39), 0.193 (0.46) \\\addlinespace[2pt]
Walking data, 1e-4, one-step windows, seed 0 & 20 & failed & 0.150 & 0.209 (0.67), 0.103 (0.75), 0.139 (0.43) \\\addlinespace[2pt]
Walking data, 1e-4, one-step windows, seed 0 & 80 & failed & 0.097 & 0.091 (0.78), 0.102 (0.86) \\\addlinespace[2pt]
Mixed data, 1e-4, rollout windows, seed 0 & 5 & failed & 0.110 & 0.000 (1.00), 0.198 (0.76), 0.132 (0.49) \\\addlinespace[2pt]
Mixed data, 1e-4, rollout windows, seed 0 & 20 & failed & 0.040 & 0.000 (1.00), 0.081 (0.30) \\\addlinespace[2pt]
Mixed data, 1e-4, rollout windows, seed 0 & 80 & failed & 0.080 & 0.081 (0.06), 0.078 (0.00) \\\addlinespace[2pt]
Walking data, 1e-5, rollout windows, seed 0 & 80 & did not fail & 0.305 & 0.181 (0.68), 0.430 (0.33) \\\addlinespace[2pt]
Walking data, 1e-5, rollout windows, seed 0 & 160 & failed & 0.088 & 0.091 (0.80), 0.085 (0.86) \\\addlinespace[2pt]
Walking data, 1e-5, rollout windows, seed 0 & 320 & failed & 0.047 & 0.047 (0.83), 0.046 (0.80) \\\addlinespace[2pt]
Mixed data, 1e-5, rollout windows, seed 0 & 80 & did not fail & 0.304 & 0.499 (0.42), 0.000 (1.00), 0.413 (0.54) \\\addlinespace[2pt]
Mixed data, 1e-5, rollout windows, seed 0 & 160 & failed & 0.124 & 0.157 (0.64), 0.083 (0.74), 0.133 (0.51) \\\addlinespace[2pt]
Mixed data, 1e-5, rollout windows, seed 0 & 320 & failed & 0.098 & 0.109 (0.45), 0.087 (0.90) \\
\end{xltabular}
\endgroup

The released model's two registered policies, for comparison: 0.610 (0.35) and 0.573 (0.31). The mean walking score of the simulator fine-tuned on one-step windows for 20 steps rounds to the threshold, yet the simulator failed, because two of its three policies scored below 0.15.

\end{document}